\documentclass[dvipsnames]{fairmeta}

\title{Let RGB Be the Language of Vision}
\author[1]{Timing Yang} \author[2]{Jinrui Yang} \author[1*]{Xinlong Li} \author[2]{Yuhan Wang} \author[3]{Haoran Li} \author[2]{Yanqing Liu}  \author[1]{Guoyizhe Wei} \author[1*]{Jixuan Ying} \author[4]{Chen Wei} \author[1]{Rama Chellappa} \author[2]{Yuyin Zhou} \author[2]{Cihang Xie} \author[1]{Alan Yuille} \author[1\dagger]{Feng Wang}
\makeatletter
\newcommand\affiliationNL[2][]{\addtolist[#1]{#2}{\affiliationlist}{\affiliationformat}{\\}}
\makeatother

\affiliation[1]{Johns Hopkins University}
\affiliation[2]{UC Santa Cruz}
\affiliation[3]{Carnegie Mellon University}
\affiliation[4]{Rice University}

\affiliationNL[\dagger]{Project lead}
\affiliation[*]{visiting students}
\date{July 12th, 2026}

\abstract{This work introduces a unified formulation for vision models, where diverse forms of visual information beyond natural images, such as masks, depth maps, and other structured visual signals, are all represented as RGB images, while general visual tasks can be converted into a common RGB-to-RGB image editing problem. In this paradigm, different types of visual information internally share the same encoding and decoding architecture and parameters as natural images, enabling a single model to transfer across tasks through a unified visual interface, in a way analogous to how language models operate over text. We refer to this formulation as RGB In and RGB Out (\textbf{RINO}). Built upon a generic image editing backbone without task-specific fine-tuning, RINO demonstrates robust and competitive zero-shot performance on both dense understanding tasks such as segmentation and depth estimation (where we unify outputs as RGB), and dense-conditioned generation tasks such as pose-to-image generation (where we unify inputs as RGB). We hope this study provides useful insights toward general unified vision-language systems, where diverse visual tasks can be expressed, interpreted, and solved through a shared visual language. Code is available at \url{https://github.com/yangtiming/RINO}.
}
\usepackage{tabularx}
\begin{document}
\maketitle

\section{Introduction}

In Large Language Models (LLMs), text serves as the universal input and output format for almost all tasks, allowing a well-trained language model to flexibly support diverse downstream applications. In vision, however, this level of unification is still far from being achieved. Different types of visual information typically require their own specific representations, such as RGB for natural images, one-hot masks for segmentation, and continuous geometric maps for depth. This diversity in representation leads to a direct challenge: for each format, we often need to design a dedicated encoder and decoder, together with additional adapter components to connect them. Such barriers in both model structure and information representation make it difficult for vision models to support free-form interactions in the same way as language models, and often limit their ability to effectively generalize beyond individual tasks, leaving most vision foundation models still closer to task-specific experts than truly general-purpose visual learners~\citep{sam,depthanything,stablediffusion,clip,dinov2}.

This naturally raises a question: is there a way for visual information to communicate as freely as text under a unified form? In other words, \textbf{\textit{can vision develop its own universal interface?}} Interestingly, recent studies provide useful insights. Vision Banana~\citep{visionbanana} proposes to solve understanding problems such as segmentation, depth estimation, and surface normal estimation with a generative model; for example, instead of predicting a segmentation mask in its conventional format, it generates the mask in RGB space based on the input image. This generative understanding paradigm has also been further verified on open-source image models and shown strong robustness~\citep{tencentbanana}. Inspired by these observations, we ask whether RGB can be used to unify both the input and output forms of visual information, serving a role similar to text in LLMs and forming a universal “visual language” that humans can naturally interact with.

To this end, we propose \textbf{RGB In and RGB Out (RINO)}, a unified visual representation learning paradigm that expresses all visual input and output signals in RGB format. Unlike Vision Banana that evaluates on specific understanding tasks, we aim to investigate how broadly this paradigm can generalize: we evaluate RINO on over 20 vision tasks/benchmarks across multiple domains, including dense understanding, 3D estimation, and conditioned generation, where all visual inputs and outputs are represented in RGB format and share the same encoding/decoding system as natural images. We build RINO upon pretrained image editing models such as Qwen-Image-Edit~\citep{qwenimage} without introducing any auxiliary parameters. For each type of visual information beyond natural images, we only employ lightweight, parameter-free data conversion modules to transform signals between RGB and other formats such as segmentation masks and depth maps, while inside the model, all such visual information is interpreted or generated purely as images.

By constructing this standard RGB/text-to-RGB/text framework, we surprisingly find that RINO can broadly handle all tested understanding and generation tasks in a zero-shot manner. It stably produces visually meaningful results and achieves quantitative performance comparable to in-domain expert models across different applications. For example, on depth estimation, RINO achieves an $\delta_1$ score of 0.938, which approaches Depth Anything~\citep{depthanything}, a specialist model trained specifically for depth estimation. We further find that for conditioned image generation, which has traditionally relied on ControlNet-like~\citep{controlnet} methods with input-specific encoders and adapters, RINO establishes a new zero-shot state of the art under existing evaluation protocols. Visually, RINO’s understanding and generation results also receive high scores in human preference studies, suggesting that RGB-based unification has strong potential for human interaction and may serve as a transferable “language” for vision.

Overall, this work provides strong evidence for the feasibility of RGB as a unified interface for vision. We present an initial attempt toward building a large-scale, multi-task unified vision-language system, and observe highly promising signs of broad task transfer under a single RGB-based formulation. At this stage, our RINO models still have limitations: their strong transfer ability relies on generic image editing backbones like Qwen-Image-Edit~\citep{qwenimage} and FireRed~\citep{firered}, whose web-scale pretraining data may already contain diverse visual signals such as segmentation masks and depth maps. However, such signals likely occupy only a very small portion of the pretraining distribution, which may lead to a performance gap between RINO and in-domain expert models on some specific applications. We refer to the current fully zero-shot version as \textbf{RINO-Zero}. In future work, we will instruction-tune existing image editing models with standard image editing data mixed with a proportion of RGB-formatted non-natural visual data, such as segmentation masks and depth maps, so that the RINO framework is introduced during training across a wider range of tasks. We hope this work can provide useful insights for future unified vision-language systems and help extend this paradigm to broader vision domains, including 3D vision, video, and world models.

\begin{figure}[t]
    \centering
    \includegraphics[width=\textwidth]{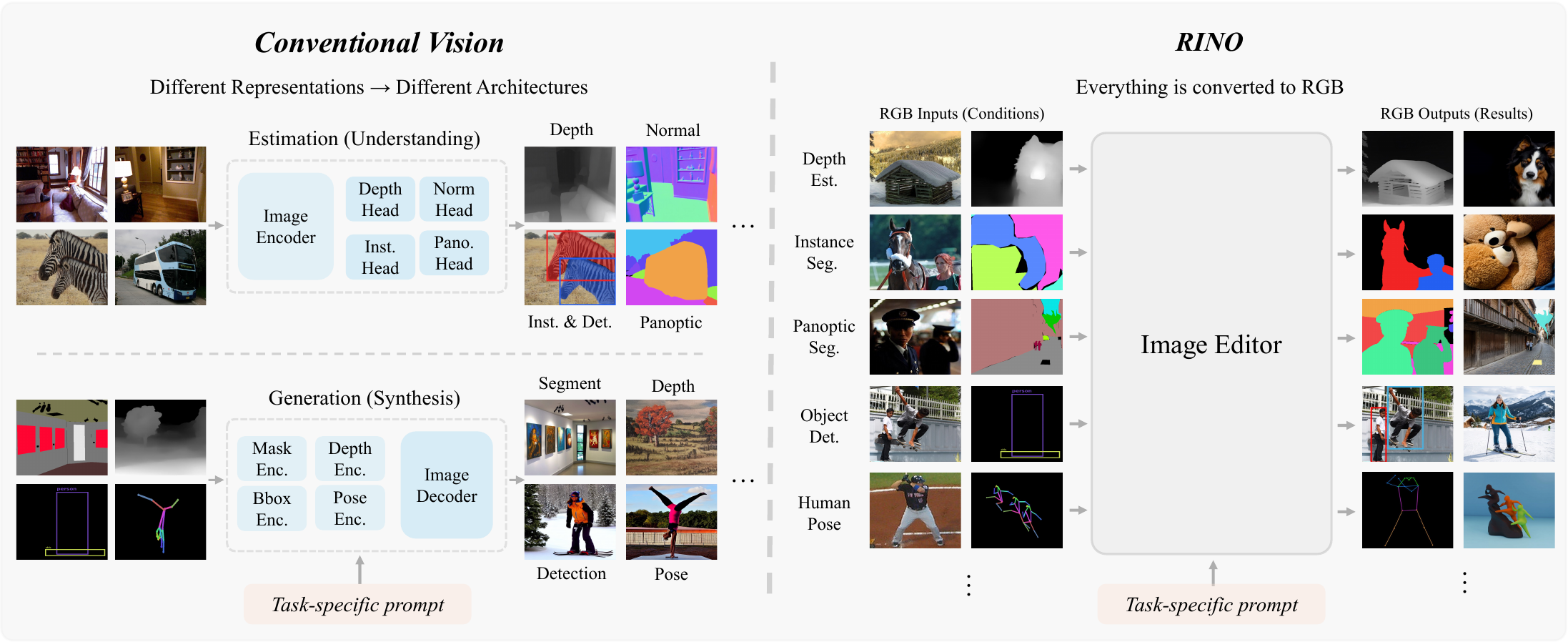}
    \caption{\textbf{RINO unifies vision under a single RGB interface.}
    \emph{Left:} conventional vision ties each task to its own representation and
    architecture, with task-specific encoders and heads or generators.
    \emph{Right:} RINO renders every input and output as RGB, so one frozen image
editor, driven by a task-specific prompt, handles estimation (depth, normals,
segmentation, detection, pose, referring) and the corresponding conditioned
generation, all without any task-specific module.}
    \label{fig:teaser}
\end{figure}

\section{Method: Unified RGB Interface for Vision}

We formulate general vision-language understanding and generation tasks under a unified RGB/text-to-RGB/text paradigm. Given an input image $x_{\mathrm{img}}$ and an input text prompt $x_{\mathrm{text}}$, a multimodal image editing model $F$ produces an output image $y_{\mathrm{img}}$ and, optionally, an output text response $y_{\mathrm{text}}$:
\begin{equation}
    y_{\mathrm{img}}, y_{\mathrm{text}} = \mathcal{F}(x_{\mathrm{img}}, x_{\mathrm{text}}),
\end{equation}
where $\mathcal{F}$ is a generic pretrained image editing model such as Qwen-Image-Edit~\citep{qwenimage}. In our framework, we do not introduce task-specific heads, encoders, decoders, or adapter parameters. Instead, all visual information is represented in RGB format and processed by the same image encoder and decoder inside $\mathcal{F}$. This allows natural images, segmentation masks, depth maps, pose maps, edge maps, and other structured visual signals to share the same visual interface. We illustrate our overall framework in~\Cref{fig:teaser}.

\textbf{For visual understanding tasks}, the input $x_{\mathrm{img}}$ is a natural image, and the input $x_{\mathrm{text}}$ specifies the task to be performed. The output image $y_{\mathrm{img}}$ is an RGB representation of the target structured visual information. For example, in semantic segmentation, $y_{\mathrm{img}}$ is an RGB segmentation mask; in depth estimation, $y_{\mathrm{img}}$ is an RGB-formatted depth map. The optional text output $y_{\mathrm{text}}$ can be used to explain the prediction or describe the generated visual result. When evaluating $y_{\mathrm{img}}$, we use a parameter-free converter to transform the RGB output into the corresponding task-specific format. For example, an RGB segmentation map can be converted into class labels through a predefined color mapping, and an RGB depth visualization can be converted into a depth map through a predefined decoding rule. These converters are only used before or after model inference. All visual reasoning is performed inside the base model $\mathcal{F}$ through the unified RGB interface.

\textbf{For visual generation tasks}, the input $x_{\mathrm{img}}$ can be an RGB-formatted visual condition, such as a semantic layout, a pose skeleton, a depth map, an edge map, or a mask. The text prompt $x_{\mathrm{text}}$ describes the desired generation or editing behavior. The model then generates an output image $y_{\mathrm{img}}$, which is usually a natural image that follows the given visual condition and text instruction. This formulation treats conditioned generation as the dual problem of visual understanding. In understanding, the model maps a natural image to an RGB-formatted structured prediction. In generation, the model maps an RGB-formatted structured condition back to a natural image. For example, if semantic segmentation can be formulated as image-to-RGB-mask prediction, then segmentation-conditioned image generation can be formulated as RGB-mask-to-image generation. Similarly, pose estimation can be viewed as image-to-RGB-pose prediction, while pose-guided generation can be viewed as RGB-pose-to-image generation.

\textbf{Base models.} We run RINO-Zero on three open-sourced image editors: Qwen-Image-Edit~\citep{qwenimage}, LongCat-Image-Edit~\citep{longcatedit} and FireRed-Image-Edit~\citep{firered}. Qwen-Image-Edit builds on Qwen-Image, a 20B-parameter multimodal diffusion transformer (MMDiT) that couples a Qwen2.5-VL vision-language model with a VAE for image tokens. LongCat-Image-Edit is a compact 6B bilingual (Chinese-English) editor from Meituan, built on a hybrid MMDiT / single-stream DiT design with a Qwen2.5-VL text encoder and tuned for efficiency at a smaller size. FireRed-Image-Edit, from the FireRed team, extends an open-source text-to-image foundation model with a double-stream multimodal-diffusion design for high-fidelity, instruction-following edits. For all three we use the released weights as a black-box RGB-to-RGB editor, without adding, removing, or fine-tuning any layers.

\section{Experiments}

\subsection{Understanding Tasks: Unifying Outputs as RGB}

\textbf{Depth Estimation.} For depth estimation, we prompt the image editor to repaint the input image as a grayscale depth visualization, where nearby surfaces are brighter and distant surfaces are darker. We then recover a dense relative depth map from the per-pixel luminance of the generated image. In this task, we evaluate with Qwen-Image-Edit~\citep{qwenimage} and LongCat-Image-Edit~\citep{longcatedit} in a zero-shot setting with a fixed sampling configuration. Since a generative editor produces relative depth with unknown scale and offset, we follow the standard affine-invariant evaluation protocol. For each image, we fit a scale and shift to the ground-truth depth by least squares before computing $\delta_1$ and AbsRel. \Cref{tab:depth} follows the \emph{disparity}-space, or inverse-depth, convention used by MiDaS and Depth~Anything~\citep{ranftl2020midas,depthanything}, while \Cref{tab:depth_linear} follows the \emph{linear} depth-space convention used by Marigold~\citep{ke2024marigold}.

\begin{table}[t!]
\centering
\caption{Zero-shot monocular depth estimation under a disparity-space
affine-invariant protocol. Specialist scores are quoted from~\citet{depthanythingv2}
(ViT-L);  $\delta_1\uparrow$/AbsRel$\downarrow$. \textbf{Bold}:
best per metric.}
\label{tab:depth}
\setlength{\tabcolsep}{6pt}
\begin{tabular}{l cc cc cc}
\toprule
& \multicolumn{2}{c}{NYUv2} & \multicolumn{2}{c}{KITTI} & \multicolumn{2}{c}{DIODE-indoor} \\
\cmidrule(lr){2-3}\cmidrule(lr){4-5}\cmidrule(lr){6-7}
Method & $\delta_1$ & AbsRel & $\delta_1$ & AbsRel & $\delta_1$ & AbsRel \\
\midrule
\multicolumn{7}{l}{\emph{Relative-depth specialists}}\\
MiDaS V3.1~\citep{midasv31}        & 0.980 & 0.048 & 0.850 & 0.127 & 0.942 & 0.075 \\
Depth Anything V1~\citep{depthanything} & \textbf{0.981} & \textbf{0.043} & \textbf{0.947} & 0.076 & \textbf{0.952} & \textbf{0.066} \\
Depth Anything V2~\citep{depthanythingv2} & 0.979 & 0.045 & 0.946 & \textbf{0.074} & \textbf{0.952} & \textbf{0.066} \\
\midrule
\multicolumn{7}{l}{\emph{Generative image editors}}\\
Qwen-Image-Edit & 0.927 & 0.086 & 0.901 & 0.093 & 0.938 & 0.076 \\
LongCat-Image-Edit   & 0.840 & 0.130 & 0.849 & 0.130 & 0.849 & 0.129 \\
\bottomrule
\end{tabular}
\end{table}

\begin{table}[t!]
\centering
\caption{The same editors under the \emph{linear} depth-space affine-invariant
protocol of~\citet{tencentbanana} (per-image least-squares scale\,+\,shift on GT
depth). $^{\dagger}$Vision~Banana~\citep{visionbanana}
reports \emph{raw metric} depth with no alignment, shown only as a reference and not
directly comparable. $\delta_1\uparrow$/AbsRel$\downarrow$. \textbf{Bold}: best editor
per metric.}
\label{tab:depth_linear}
\setlength{\tabcolsep}{6pt}
\begin{tabular}{l cc cc cc cc}
\toprule
& \multicolumn{2}{c}{NYUv2} & \multicolumn{2}{c}{KITTI} & \multicolumn{2}{c}{DIODE-indoor} & \multicolumn{2}{c}{iBims-1} \\
\cmidrule(lr){2-3}\cmidrule(lr){4-5}\cmidrule(lr){6-7}\cmidrule(lr){8-9}
Method & $\delta_1$ & AbsRel & $\delta_1$ & AbsRel & $\delta_1$ & AbsRel & $\delta_1$ & AbsRel \\
\midrule
\multicolumn{9}{l}{\emph{Metric reference (instruction-tuned)}}\\
Vision Banana$^{\dagger}$ & 0.948 & 0.081 & 0.915 & 0.107 & 0.917 & 0.108 & 0.934 & 0.078 \\
\midrule
\multicolumn{9}{l}{\emph{Generative image editors (linear affine-invariant)}}\\
Qwen-Image-Edit & 0.792 & 0.156 & 0.413 & 0.329 & \textbf{0.872} & \textbf{0.122} & 0.824 & 0.133 \\
LongCat-Image-Edit   & \textbf{0.832} & \textbf{0.135} & \textbf{0.454} & \textbf{0.325} & 0.825 & 0.146 & \textbf{0.843} & \textbf{0.129} \\
\bottomrule
\end{tabular}
\end{table}

\begin{figure}[t!]
  \centering
  \includegraphics[width=\linewidth]{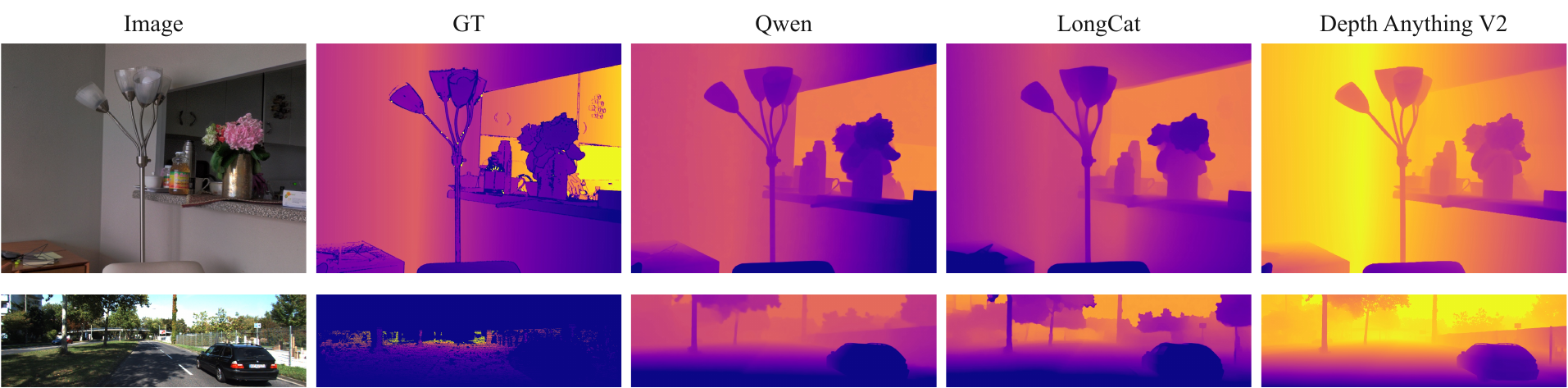}
  \caption{Qualitative zero-shot depth estimation on DIODE-indoor (top) and KITTI (bottom).}
  \label{fig:vis_depth}
\end{figure}

\Cref{tab:depth} compares the two editors with three relative-depth specialists in disparity space: MiDaS~V3.1~\citep{midasv31}, Depth~Anything~V1~\citep{depthanything}, and Depth~Anything~V2~\citep{depthanythingv2}, with specialist results quoted from~\citet{depthanythingv2}. Although neither editor is trained for depth estimation or equipped with any depth-specific parameters, Qwen-Image-Edit performs surprisingly close to dedicated models. It remains within a few $\delta_1$ points of the specialists on DIODE-indoor~\citep{vasiljevic2019diode} ($0.938$ vs.\ $0.952$), and shows a roughly five-point gap on NYUv2~\citep{silberman2012indoor} and KITTI~\citep{geiger2012kitti}. \Cref{tab:depth_linear} reports the same comparison in linear depth space, where performance drops on wide-range scenes such as KITTI and iBims-1~\citep{koch2018evaluation}. We include Vision~Banana~\citep{visionbanana} as a reference in this table, since it predicts raw metric depth without per-image alignment and is therefore not directly comparable to the affine-invariant editor outputs. Qualitative visualizations are shown in \Cref{fig:vis_depth}. We observe that RINO's prediction can clearly reflect the depth layout of input images.

\begin{table}[t]
\centering
\caption{Zero-shot surface-normal estimation on NYUv2, iBims-1, and DIODE-indoor. We report Mean$\downarrow$ and median$\downarrow$ angular error (degrees). \textbf{Bold}: best editor per metric.}
\label{tab:normals}
\setlength{\tabcolsep}{6pt}
\begin{tabular}{l cc cc cc}
\toprule
& \multicolumn{2}{c}{NYUv2} & \multicolumn{2}{c}{iBims-1} & \multicolumn{2}{c}{DIODE-indoor} \\
\cmidrule(lr){2-3}\cmidrule(lr){4-5}\cmidrule(lr){6-7}
Method & Mean & Median & Mean & Median & Mean & Median \\
\midrule
\multicolumn{7}{l}{\emph{Surface-normal specialists (quoted; see caption)}}\\
DSINE~\citep{DSINE}                     & 16.4  & 8.4   & 17.1  & 6.1  & 18.45 & 13.87 \\
Marigold~\citep{ke2024marigold}         & 20.86 & 11.13 & 18.46 & 8.44 & 16.67 & 12.08 \\
StableNormal~\citep{ye2024stablenormal} & 19.71 & 10.53 & 17.25 & 8.06 & 13.70 & 9.46  \\
Lotus-2~\citep{lotus2}                  & 16.9  & ---   & 15.4  & ---  & 18.58 & ---   \\
Vision~Banana~\citep{visionbanana}      & 17.78 & 8.88  & ---   & ---  & 13.82 & 11.56 \\
\midrule
\multicolumn{7}{l}{\emph{Generative image editors (ours, zero-shot, RGB$\to$RGB)}}\\
Qwen-Image-Edit   & 20.21 & 13.46 & 20.66 & 12.03 & 21.99 & 18.14 \\
FireRed-Image-Edit& \textbf{17.73} & \textbf{10.99} & \textbf{18.62} & \textbf{10.38} & \textbf{17.25} & \textbf{13.64} \\
\bottomrule
\end{tabular}
\end{table}

\begin{figure}[t!]
  \centering
  \includegraphics[width=\linewidth]{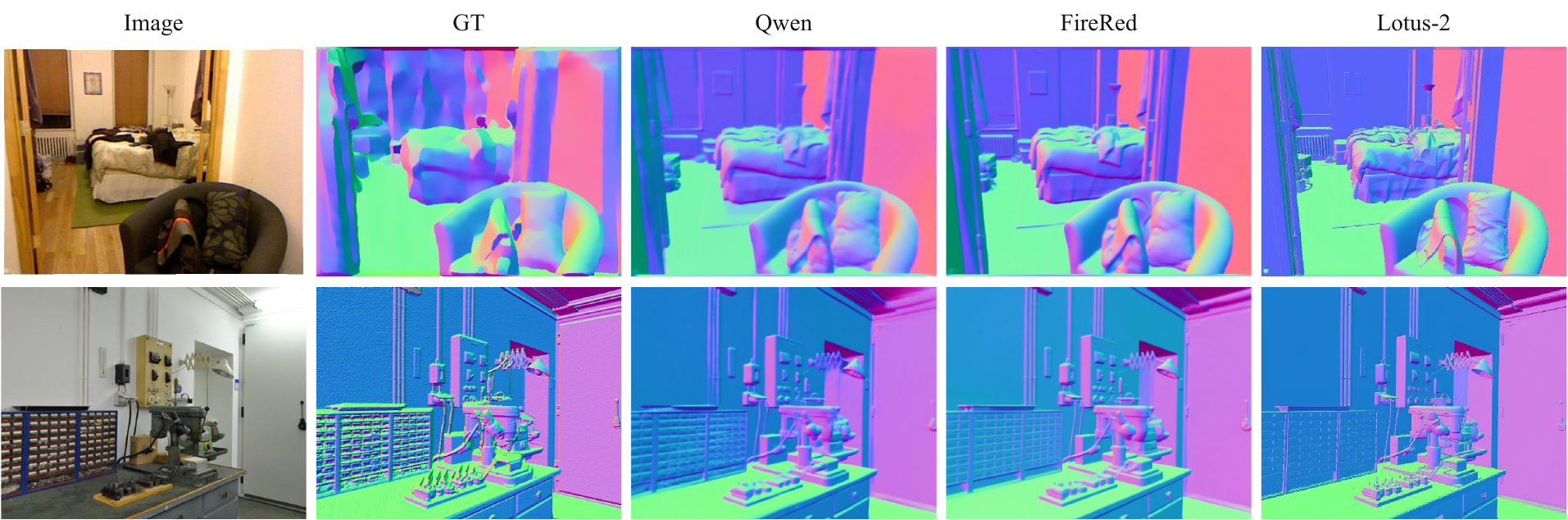}
  \caption{Qualitative zero-shot surface normal estimation on NYUv2 (top) and iBims-1 (bottom).}
  \label{fig:vis_normal}
\end{figure}

\textbf{Surface Normal Estimation.} Surface normal estimation follows the same RGB-to-RGB paradigm as depth estimation. We prompt the editors to repaint the input scene as a standard normal map, where surface orientation is encoded by color, and decode the generated RGB image back into per-pixel unit normals. We evaluate Qwen-Image-Edit and FireRed-Image-Edit in a zero-shot setting. The image editors output normals in an unknown coordinate convention, which we resolve once following~\citet{tencentbanana}, and report the standard mean and median angular error (degrees). DSINE~\citep{DSINE} is scored through the same pipeline as a specialist reference. \Cref{tab:normals} reports surface normal estimation results on NYUv2, iBims-1, and DIODE-indoor. FireRed, a generic image editor without normal-specific training or parameters, is the strongest editor across all three datasets. It only slightly trails the DSINE specialist in mean angular error, although DSINE remains sharper at the pixel level, as reflected by its lower median error. These results suggest that a zero-shot image editor can approach a dedicated normal estimation model in average surface orientation prediction, while still leaving room for improvement in local precision. Qualitative visualizations are shown in \Cref{fig:vis_normal}.

\textbf{Semantic Segmentation.} We follow ~\citet{tencentbanana}, where only the categories present in the ground-truth of the current image are included in the prompt. To mitigate common generation artifacts such as color jitter and noisy pixels, we apply a lightweight parameter-free refinement stage after color decoding, consisting of local majority filtering and small connected-component removal. To handle the large label space of ADE20K, we introduce a hierarchical segmentation strategy: the model first predicts masks for 10 super-class, and fine-grained categories are subsequently generated within their corresponding regions. Unless otherwise specified, all image editing backbones use 12 denoising steps, CFG = 5.0, and the same refinement pipeline. LongCat-Image-Edit employs 50 denoising steps due to its lower generation stability. We use three standard segmentation metrics: mean Intersection over Union (mIoU), mean class accuracy (mAcc), and all-pixel accuracy (aAcc). Quantatitive results are reported on the validation sets of Cityscapes~\citep{cordts2016cityscapes}, Pascal VOC ~\citep{everingham2010pascal}, and ADE20K~\citep{ade20k}.

\begin{table}[t!]
\centering
\caption{Semantic segmentation results on Cityscapes and Pascal VOC. \textbf{Bold} and \underline{underline} indicate the best and second-best results among generative image editors.}
\label{tab:segmentation_main}
\setlength{\tabcolsep}{3pt}
\begin{tabularx}{\textwidth}{X ccc ccc}
\toprule
& \multicolumn{3}{c}{Pascal VOC} & \multicolumn{3}{c}{Cityscapes} \\
\cmidrule(lr){2-4} \cmidrule(lr){5-7}
Method & mIoU $\uparrow$ & mAcc $\uparrow$ & aAcc $\uparrow$ & mIoU $\uparrow$ & mAcc $\uparrow$ & aAcc $\uparrow$ \\
\midrule
\multicolumn{7}{l}{\emph{Zero-shot Segmentation}}\\
FC-CLIP~\citep{yu2023convolutions} & --- & --- & --- & 56.29 & 65.42 & 78.46 \\
MaskCLIP~\citep{dong2023maskclip,barsellotti2024training} & 38.85 & 45.21 & 71.63 & --- & --- & --- \\
GroupViT~\citep{xu2022groupvit} & 52.37 & 56.97 & 79.28 & --- & --- & --- \\
OVSegmentor~\citep{xu2023learning} & 53.82 & 58.36 & 80.51 & --- & --- & --- \\
\midrule
\multicolumn{7}{l}{\emph{Generative image editors (zero-shot, RGB$\rightarrow$RGB)}}\\

Vision Banana~\citep{visionbanana} & {\color{gray}---} & {\color{gray}---} & {\color{gray}---} & {\color{gray}69.90} & {\color{gray}---} & {\color{gray}---} \\
Direct Editing Baseline~\citep{tencentbanana} & --- & --- & --- & 23.02 & 42.63 & \underline{64.14} \\
Ours (Qwen-Image-Edit) & 44.69 & 48.00 & \textbf{85.17} & \underline{24.30} & \textbf{44.38} & \textbf{66.59} \\
Ours (FireRed-Image-Edit) & \underline{45.17} & \underline{61.47} & 80.82 & 21.11 & \underline{42.70} & 58.00 \\
Ours (LongCat-Image-Edit) & \textbf{49.68} & \textbf{66.90} & \underline{83.53} & 13.42 & 26.72 & 49.23 \\
\bottomrule
\end{tabularx}
\end{table}

\begin{table}[t!]
\centering
\caption{Semantic segmentation results on ADE20K under both the 10-class coarse setting and the standard 150-category setting. \textbf{Bold} and \underline{underline} indicate the best and second-best results among generative image editors.}
\label{tab:segmentation_ade20k}
\setlength{\tabcolsep}{4pt}
\begin{tabularx}{\textwidth}{X ccc ccc}
\toprule
& \multicolumn{3}{c}{ADE20K (10 classes)} & \multicolumn{3}{c}{ADE20K (150 categories)} \\
\cmidrule(lr){2-4}\cmidrule(lr){5-7}
Method & mIoU $\uparrow$ & mAcc $\uparrow$ & aAcc $\uparrow$ & mIoU $\uparrow$ & mAcc $\uparrow$ & aAcc $\uparrow$ \\
\midrule
\multicolumn{7}{l}{\emph{Zero-shot Segmentation}}\\
MaskCLIP~\citep{dong2023maskclip} & 39.26 & 50.52 & 65.66 & 23.78 & 31.54 & 50.27 \\
FC-CLIP~\citep{yu2023convolutions} & 42.17 & 53.11 & 67.05 & 34.13 & 40.86 & 59.62 \\
\midrule
\multicolumn{7}{l}{\emph{Generative image editors (ours, zero-shot, RGB$\rightarrow$RGB)}}\\
Qwen-Image-Edit
& \textbf{40.37} & \textbf{53.51} & \textbf{67.35}
& \underline{12.57} & \underline{20.93} & \textbf{44.85} \\
FireRed-Image-Edit
& \underline{34.76} & \underline{46.79} & \underline{54.77}
& \textbf{13.61} & \textbf{22.80} & \underline{38.95} \\
LongCat-Image-Edit
& 19.33 & 30.02 & 26.37
& 7.43 & 14.60 & 26.79 \\
\bottomrule
\end{tabularx}
\end{table}

\begin{figure}[t!]
\centering
\includegraphics[width=\linewidth]{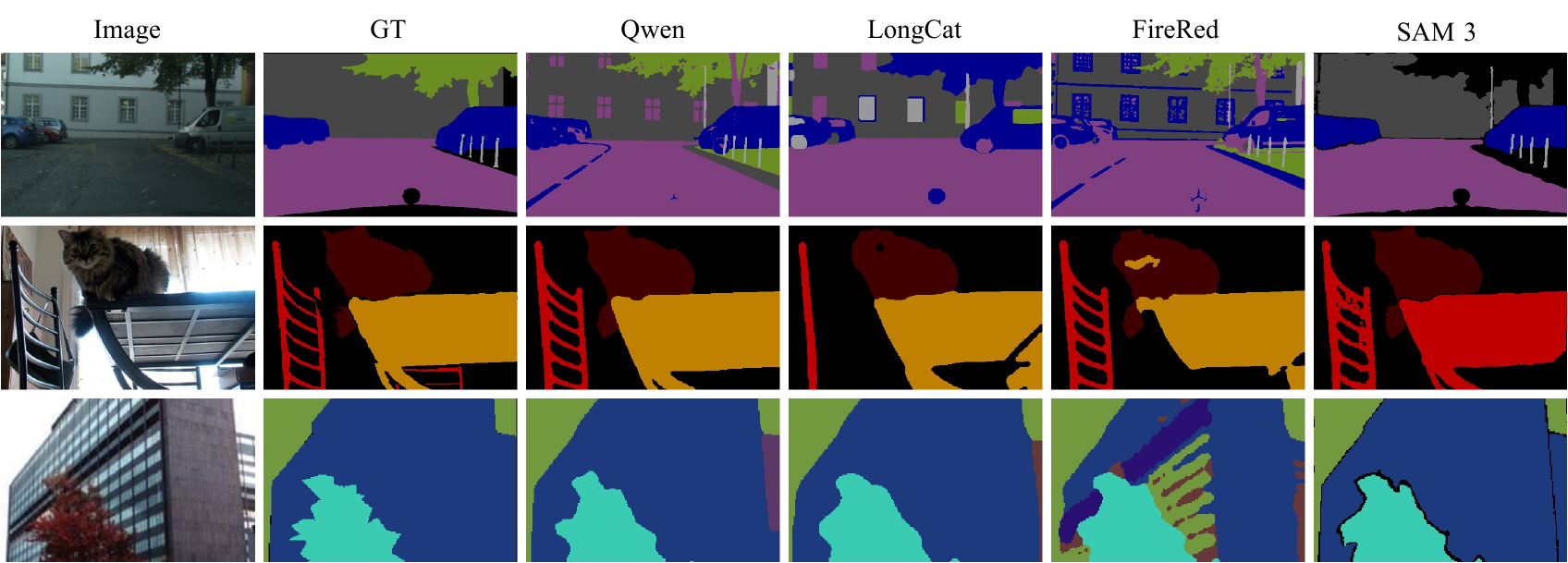}
\caption{Zero-shot semantic segmentation on Cityscapes (top), Pascal VOC (middle), and ADE20K (bottom).}
\label{fig:seg_qualitative}
\end{figure}

\Cref{tab:segmentation_main} shows that RINO achieves semantic segmentation performance comparable to zero-shot specialist models, with mAcc and aAcc even surpassing segmentation experts on PASCAL VOC. \Cref{tab:segmentation_ade20k} further evaluates segmentation at both coarse and fine granularities on ADE20K. We observe that under the coarse setting with 10 classes, RINO can even outperform the semantic segmentation specialist MaskCLIP. These results suggest that RINO has strong potential for dense prediction: the model already captures rich image semantics, but may require more domain-specific knowledge to better align its predictions with fine-grained category names, such as the 150 classes in ADE20K. Visualization results are shown in \Cref{fig:seg_qualitative}.

\begin{table}[t!]
\centering
\caption{Object detection and instance segmentation on COCO (box and mask AP, \%).
Editors are run by us zero-shot under an oracle-class, oracle-count per-class
silhouette protocol, the same blobs giving boxes and masks (all-points AP,
macro-averaged over 80 classes). 
\textbf{Bold}: best editor.}
\label{tab:detection}
\setlength{\tabcolsep}{8pt}
\begin{tabular}{l cc cc}
\toprule
 & \multicolumn{2}{c}{box AP} & \multicolumn{2}{c}{mask AP} \\
\cmidrule(lr){2-3}\cmidrule(lr){4-5}
Method & AP$_{50}$ & AP & AP$_{50}$ & AP \\
\midrule
\multicolumn{5}{l}{\emph{Supervised detect + segment models (AP quoted from their papers)}}\\
Mask DINO~\citep{maskdino}             & ---  & 50.5 & ---  & 46.0 \\
Cascade Mask R-CNN~\citep{cascadercnn} & 61.7 & 43.3 & 58.6 & 37.1 \\
Mask R-CNN~\citep{maskrcnn}            & 63.5 & 42.8 & 60.5 & 38.3 \\
Mask2Former~\citep{mask2former}            & 69.4 & 49.3 & 69.4 & 46.1 \\
\midrule
\multicolumn{5}{l}{\emph{Generative image editors (ours, zero-shot, RGB$\to$RGB)}}\\
Qwen-Image-Edit   & \textbf{16.3} & \textbf{9.4} & \textbf{13.9} & \textbf{7.2} \\
FireRed-Image-Edit& 14.9 & 8.9 & 12.9 & 6.9 \\
LongCat-Image-Edit     &  6.3 & 3.2 &  4.9 & 2.4 \\
\bottomrule
\end{tabular}
\end{table}

\begin{figure}[t!]
  \centering
  \includegraphics[width=\linewidth]{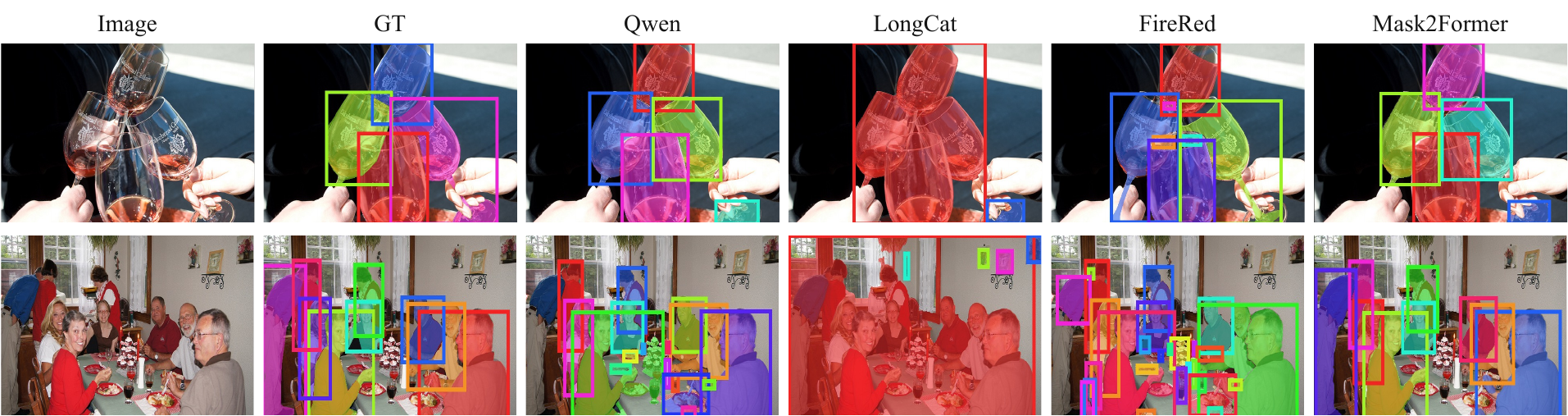}
  \caption{Qualitative object detection and instance segmentation on COCO val2017.}
  \label{fig:vis_detection}
\end{figure}

\textbf{Object Detection and Instance Segmentation.} Object detection and instance segmentation are challenging vision tasks as they require not only recognizing per-pixel semantics but also distinguishing different instances within the same category. As a result, open-vocabulary zero-shot models often struggle on datasets such as COCO, which contains 80 object categories. In RINO, we perform both tasks through per-class silhouette painting. For each present class, we prompt the editor to paint its instances as solid white blobs on a black background, while also specifying the expected number of instances. We then threshold the output and split touching blobs using morphological opening. Each connected component is treated as one instance: the blob defines its mask, and its tight bounding box defines the detection. Thus, a single prediction can be evaluated for both object detection and instance segmentation. The queried classes and instance counts are taken from the ground truth, i.e., an oracle-class and oracle-count setting, to isolate localization ability from open-set recognition. We evaluate Qwen-Image-Edit~\citep{qwenimage}, FireRed-Image-Edit~\citep{firered}, and LongCat-Image-Edit~\citep{longcatedit} in a zero-shot manner on COCO val2017. 

As shown in \Cref{tab:detection} and \Cref{fig:vis_detection}, although these tasks are highly challenging for zero-shot models, RINO still produces visually meaningful predictions, achieving 16.3\% box AP$_{50}$ and 13.9\% mask AP$_{50}$. These results remain below in-domain specialist models, such as Mask R-CNN with 63.5\% box AP$_{50}$, but they demonstrate that a zero-shot image editing model can already reach a non-trivial level of localization and instance-level prediction on COCO. \Cref{fig:vis_detection} further suggests that the relatively low quantitative scores are often caused by task-specific failure modes such as over-detection or over-segmentation, as observed in the Qwen and FireRed examples. We believe these issues can be substantially mitigated with a small amount of task-specific fine-tuning, and expect future versions of RINO to match the performance of specialist models.

\begin{table}[t]
\centering
\caption{Zero-shot panoptic segmentation on the Cityscapes, ADE20K and COCO
validation sets. Supervised specialists report Panoptic Quality (PQ, \%) as in their
papers; editor cells report Segmentation Quality (SQ, \%) and additionally give
{\scriptsize(PQ/RQ)} in parentheses (our measurements). \textbf{Bold}: best SQ results.}
\label{tab:panoptic}
\setlength{\tabcolsep}{6pt}
\begin{tabular}{l ccc}
\toprule
Method & Cityscapes & ADE20K & COCO \\
\midrule
\multicolumn{4}{l}{\emph{Supervised specialists}}\\
EoMT~\citep{eomt}                 & ---  & 51.7 & 58.3 \\
ViT-P~\citep{vitp}                & 70.8 & 51.9 & 58.0 \\
OneFormer~\citep{oneformer}       & 67.2 & 49.8 & 57.9 \\
Mask2Former~\citep{mask2former}   & 66.6 & 48.1 & 57.8 \\
\midrule
\multicolumn{4}{l}{\emph{Generative image editors}}\\
Qwen-Image-Edit   & \textbf{64.8}\,{\scriptsize(13.6/17.4)} & \textbf{71.5}\,{\scriptsize(12.3/15.9)} & 74.3\,{\scriptsize(11.9/15.5)} \\
FireRed-Image-Edit& 55.9\,{\scriptsize( 8.7/11.2)} & 67.2\,{\scriptsize( 9.4/12.5)} & 70.7\,{\scriptsize( 9.1/12.1)} \\
LongCat-Image-Edit     & 55.4\,{\scriptsize( 1.0/ 1.4)}  & 67.6\,{\scriptsize( 6.7/ 8.4)}  & \textbf{75.2}\,{\scriptsize( 6.9/ 8.8)}  \\
\bottomrule
\end{tabular}
\end{table}

\begin{figure}[t!]
  \centering
  \includegraphics[width=\linewidth]{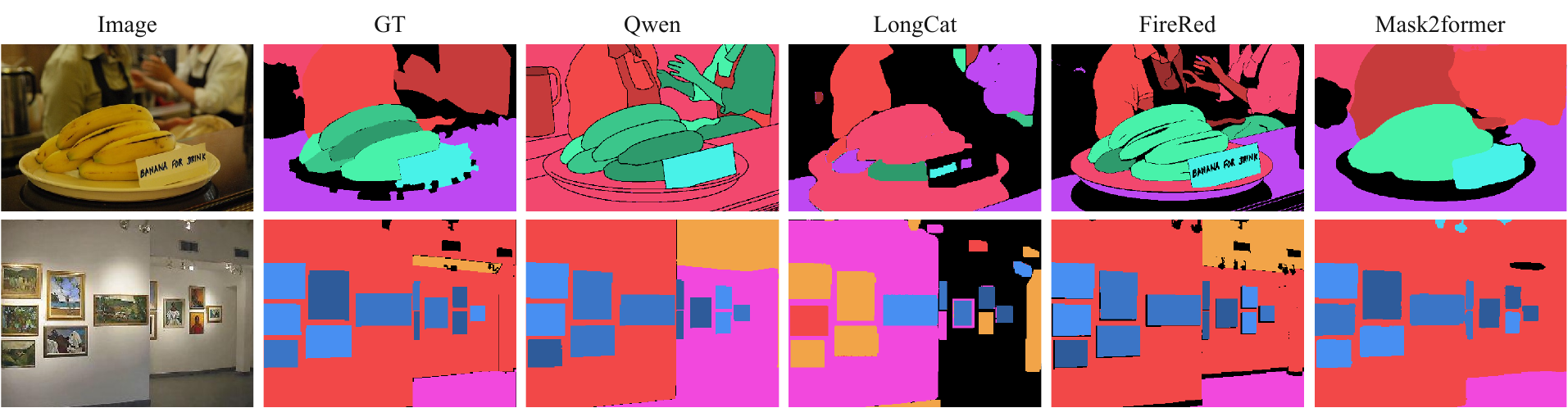}
  \caption{Qualitative zero-shot panoptic segmentation on COCO (top) and ADE20K (bottom).}
  \label{fig:vis_panoptic}
\end{figure}

\textbf{Panoptic Segmentation.} We prompt the editor to paint each region with a flat color, using one named color per class and black for the background. We then decode the output by nearest-color matching: stuff classes are treated as single segments, while thing classes are split into connected components. Following the zero-shot oracle-class protocol, only the classes present in each image are specified in the prompt. We evaluate Qwen-Image-Edit~\citep{qwenimage}, FireRed-Image-Edit~\citep{firered}, and LongCat-Image-Edit~\citep{longcatedit}. We report Panoptic Quality (PQ) and its two components, Segmentation Quality (SQ) and Recognition Quality (RQ), where $\mathrm{PQ}=\mathrm{SQ}\times\mathrm{RQ}$. PQ is the standard metric for panoptic segmentation: SQ measures the mask quality of matched regions, while RQ measures recognition and matching quality.

\Cref{tab:panoptic} reports results on Cityscapes~\citep{cordts2016cityscapes}, ADE20K~\citep{ade20k}, and COCO~\citep{lin2014microsoft}. Among the editors, Qwen-Image-Edit performs best, followed by FireRed and LongCat. Notably, despite no panoptic-specific training, the editors already show strong zero-shot spatial segmentation ability. Their predicted masks are often well aligned with object and region boundaries, leading to high SQ scores (Qwen SQ $65$--$74$). The main limitation instead lies in recognition and instance association, as reflected by lower RQ scores (Qwen RQ $15$--$17$). In many cases, coherent regions are correctly separated but assigned to wrong semantic categories, or split into multiple segments due to task-specific decoding ambiguities. This issue is most pronounced for \emph{thing} classes, where adjacent instances and small objects are often merged or missed; on Cityscapes, Qwen reaches $22.4$ PQ on \emph{stuff} but only $1.5$ PQ on \emph{things}. These results suggest that RINO can already produce structurally meaningful panoptic predictions zero-shot, while the remaining gap is largely driven by task-specific recognition and instance-matching factors rather than spatial segmentation quality. This is also supported by the visualizations in \Cref{fig:vis_panoptic}.

\begin{table}[t]
\centering
\caption{Zero-shot 2D human pose estimation on COCO val2017 single-person crops.
PCK is bbox-normalized and reported both in absolute image coordinates
(\emph{raw}) and after per-image similarity alignment (\emph{PA}), which isolates
pose structure by removing scale and translation. Dedicated pose estimators are
evaluated under the same top-down protocol for direct comparison.}
\label{tab:pose}
\setlength{\tabcolsep}{10pt}
\begin{tabular}{l ccc}
\toprule
Method & raw PCK@0.05 & PA PCK@0.05 & PA PCK@0.2 \\
\midrule
\multicolumn{4}{l}{\emph{2D-pose specialists (run under our protocol)}}\\
ViTPose~\citep{vitpose}         & 0.868 & 0.889 & 0.984 \\
Keypoint R-CNN~\citep{maskrcnn} & 0.732 & 0.765 & 0.919 \\
YOLO-pose~\citep{yolopose}      & 0.730 & 0.782 & 0.953 \\
\midrule
\multicolumn{4}{l}{\emph{Mesh-recovery specialist}}\\
SAM 3D Body~\citep{sam3dbody}    & 0.868 & --- & --- \\
\midrule
\multicolumn{4}{l}{\emph{Generative image editors}}\\
LongCat-Image-Edit     & 0.233 & 0.282 & 0.767 \\
Qwen-Image-Edit   & 0.060 & 0.235 & 0.774 \\
FireRed-Image-Edit& 0.047 & 0.243 & 0.795 \\
\bottomrule
\end{tabular}
\end{table}

\begin{figure}[t!]
  \centering
  \includegraphics[width=\linewidth]{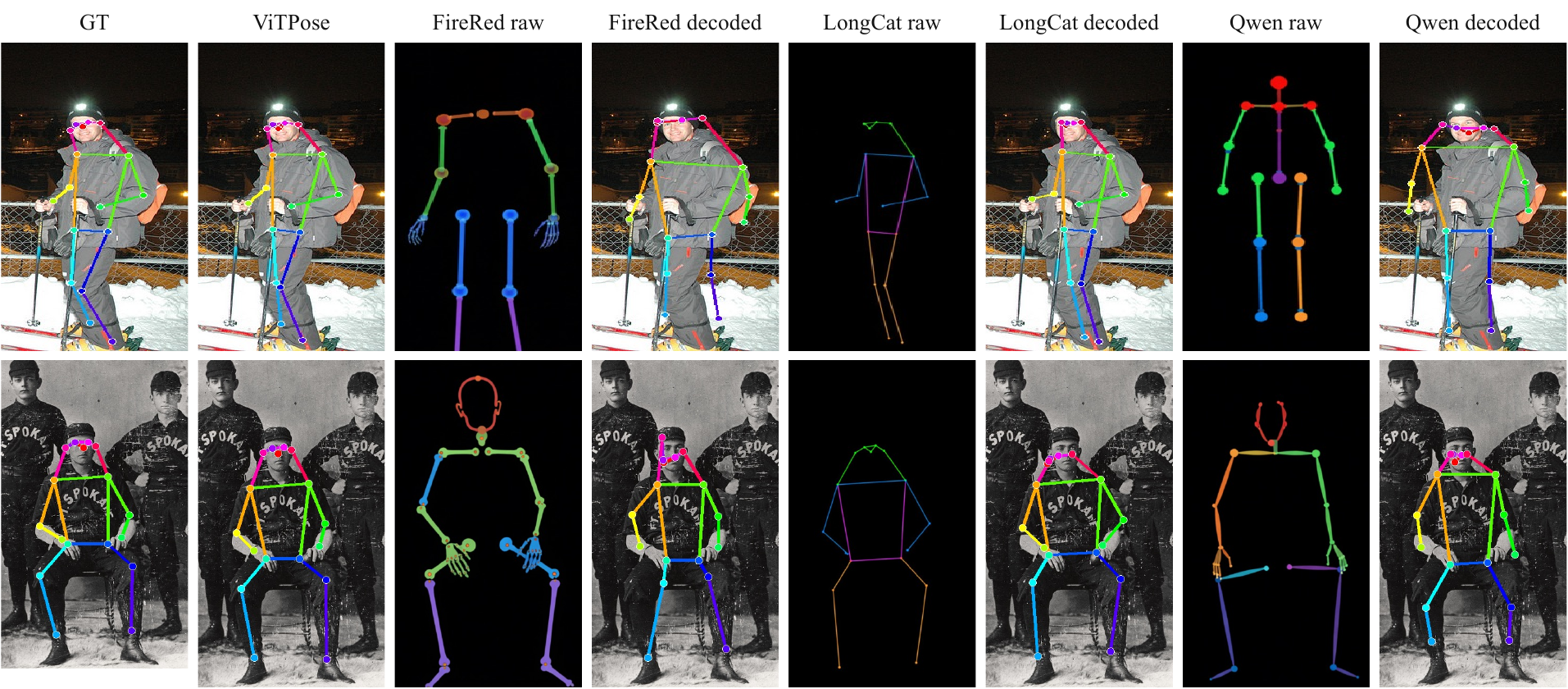}
  \caption{Qualitative zero-shot 2D human pose estimation.}
  \label{fig:vis_pose}
\end{figure}

\textbf{Human Pose Estimation.} Following the top-down protocol, each person is cropped using its ground-truth box and letterboxed to a fixed canvas. We prompt the editor to repaint the crop as an OpenPose-style~\citep{cao2021openpose} colored skeleton on a black background, and then decode COCO-17 keypoints from the output. Since image editors do not reliably reproduce a fixed color palette, we use a color-agnostic decoder: joint candidates are extracted as peaks of the foreground distance transform and labeled by matching to a canonical upright skeleton template. No keypoint head or auxiliary parameters are introduced. We evaluate Qwen-Image-Edit~\citep{qwenimage}, FireRed-Image-Edit~\citep{firered}, and LongCat-Image-Edit~\citep{longcatedit} in a zero-shot manner on COCO val2017 single-person crops. Since a generative editor may predict pose up to an unknown similarity transform, we report bbox-normalized PCK@$\tau$ under two protocols: \emph{raw}, in absolute image coordinates, following dedicated estimators, and \emph{PA}, after a per-image scale-and-translation fit to the ground truth, which isolates pose structure.

\Cref{tab:pose} reports results on COCO val2017. The editors already produce non-trivial zero-shot pose predictions, often recovering coherent human-body layouts and plausible limb structures. Although precise keypoint localization remains challenging, the generated skeletons are visually meaningful and can be decoded into COCO-17 keypoints without any pose-specific head or training. The improved performance under similarity alignment further suggests that these models capture the coarse structure of human pose, even when absolute localization is imperfect. These results indicate that RGB skeleton prediction is a viable interface for pose estimation under the RINO framework. With lightweight task-specific fine-tuning, we expect this ability to become substantially stronger and more reliable. Qualitative visualizations are shown in \Cref{fig:vis_pose}.

\begin{table}[t]
\centering
\caption{Referring expression comprehension (REC) and segmentation (RES) on RefCOCOg (UMD split) val. REC: box Prec@0.5; RES: cIoU/mIoU. \textbf{Bold}: best editor per metric.}
\label{tab:refcoco}
\setlength{\tabcolsep}{10pt}
\begin{tabular}{l c cc}
\toprule
 & REC & \multicolumn{2}{c}{RES} \\
\cmidrule(lr){2-2}\cmidrule(lr){3-4}
Method & Prec@0.5 & cIoU & mIoU \\
\midrule
\multicolumn{4}{l}{\emph{Supervised grounding specialists}}\\
LAVT~\citep{lavt} & --- & 61.2 & --- \\
UNINEXT~\citep{uninext} & 88.7 & 74.7 & ---  \\
OneRef~\citep{oneref}   & 88.1 & ---  & 73.2 \\
\midrule
\multicolumn{4}{l}{\emph{Generative image models}}\\
Vision Banana~\citep{visionbanana} & --- & 73.8 & ---  \\
Qwen-Image-Edit   & \textbf{51.8} & \textbf{40.3} & \textbf{45.3} \\
LongCat-Image-Edit     & 47.7 & 35.0 & 38.7 \\
FireRed-Image-Edit& 47.6 & 35.7 & 41.6 \\
\bottomrule
\end{tabular}
\end{table}

\begin{figure}[t!]
  \centering
  \includegraphics[width=\linewidth]{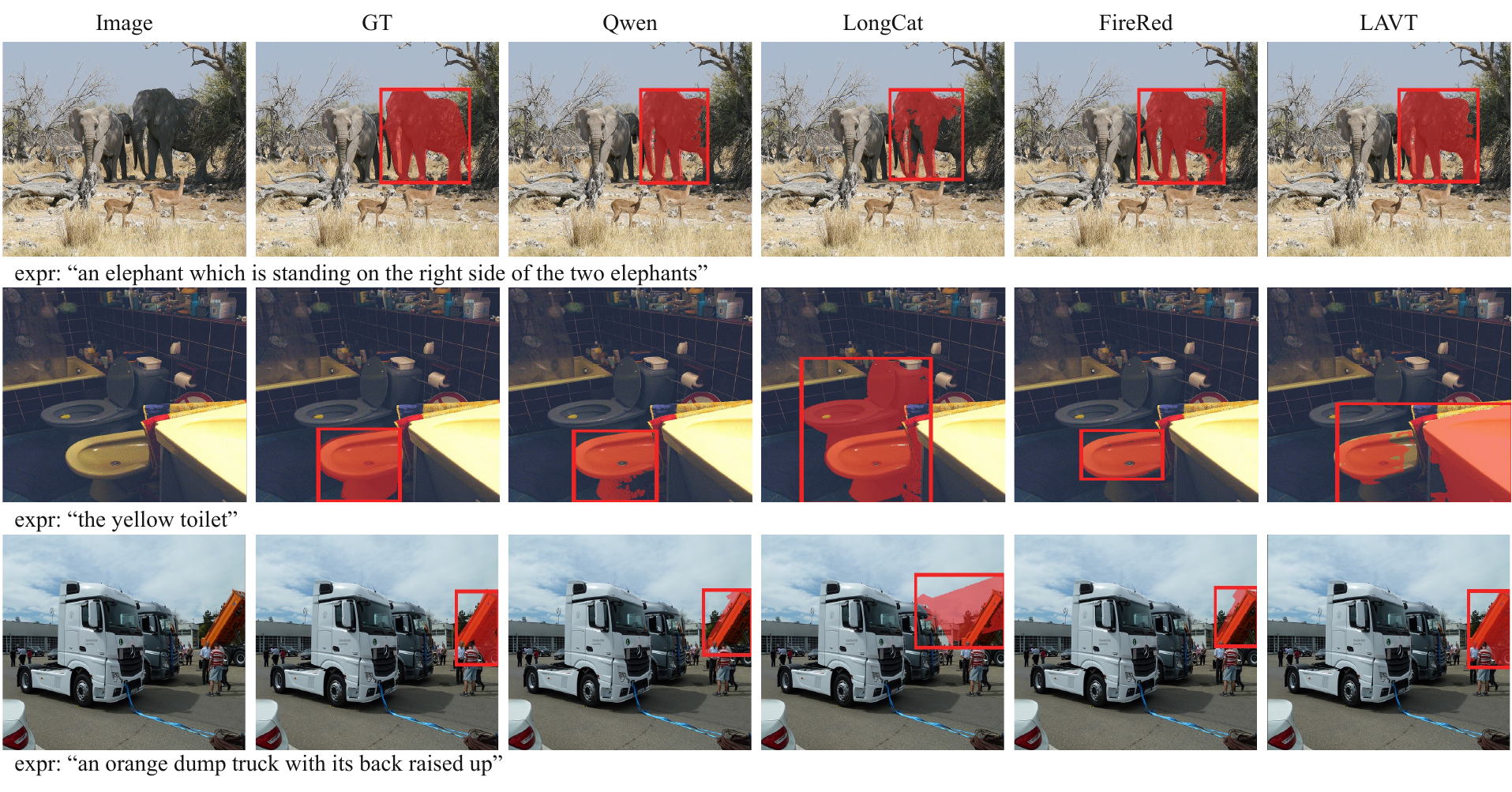}
  \caption{Qualitative referring expression comprehension and segmentation on RefCOCOg.}
  \label{fig:vis_refcoco}
\end{figure}

\textbf{Referring Expression Comprehension and Segmentation.} We perform language grounding through image editing. For each referring expression, we prompt the editor to cover the referred object in solid red while leaving all other pixels unchanged. We then compute the difference between the edited output and the input image, and keep the largest changed connected component as the prediction: its pixels define the segmentation mask, and its tight bounding box defines the comprehension result. The editor only receives the image and the expression, without access to boxes, classes, or instance counts, and no grounding head is introduced. We evaluate Qwen-Image-Edit, FireRed-Image-Edit, and LongCat-Image-Edit zero-shot on the RefCOCOg UMD validation split~\citep{mao2016generation,nagaraja2016modeling}. We report the standard metrics for both subtasks: box Prec@0.5 for referring expression comprehension (REC), where a prediction is correct if its box IoU is at least $0.5$, and cumulative IoU (cIoU) for referring expression segmentation (RES), computed as dataset-pooled intersection-over-union. Missed predictions are assigned zero score.

As shown in \Cref{tab:refcoco}, the open-source editors achieve meaningful zero-shot grounding from free-form language to image regions. Although their segmentation accuracy is still below Vision~Banana~\citep{visionbanana}, whose cIoU is comparable to supervised grounders such as UNINEXT and OneRef, they can already localize and segment referred objects without any grounding-specific head, box supervision, or task-specific fine-tuning. This is a substantially harder setting than oracle-class segmentation, since the model must jointly parse the referring expression, identify the target instance, and express the grounded region as an RGB edit. The results suggest that generic image editors already possess a non-trivial ability to connect language with pixel-level visual grounding through the RGB editing interface. Qualitative examples in \Cref{fig:vis_refcoco} further show that many predictions are visually interpretable and spatially well formed, indicating a promising direction for improving referring understanding under the RINO framework.

\begin{table}[t]
\centering
\caption{Open-vocabulary instance segmentation on the \emph{attributes} subset of SA-Co/Gold ($\mathrm{cgF1}=100\cdot\mathrm{pmF1}\cdot\mathrm{IL\_MCC}$) Note that $\mathrm{pmF1}$ is gate-independent and we obtain comparable results as specialist models. SAM~3 is supervised; Vision~Banana adds an MLLM presence gate. \textbf{Bold:} best editor results.}
\label{tab:saco}
\setlength{\tabcolsep}{10pt}
\begin{tabular}{l ccc}
\toprule
Method & pmF1 & IL\_MCC & cgF1 \\
\midrule
\multicolumn{4}{l}{\emph{Supervised specialist (quoted, not zero-shot transfer)}}\\
SAM 3~\citep{sam3}                       & 72.0 & 0.76 & 54.9 \\
\midrule
\multicolumn{4}{l}{\emph{Zero-shot transfer}}\\
gDino-T~\citep{groundingdino}        & 47.3 & 0.29 & 13.8 \\
Vision Banana + Gemini~\citep{visionbanana} & 68.7 & 0.86 & 58.8 \\
\midrule
\multicolumn{4}{l}{\emph{Generative image editors (ours, zero-shot, RGB$\to$RGB)}}\\
Qwen-Image-Edit   & \textbf{39.8} & 0.04 & \textbf{1.7} \\
FireRed-Image-Edit& 30.9 & 0.01 & 0.2 \\
LongCat-Image-Edit     & 17.3 & \textbf{0.09} & 1.6 \\
\bottomrule
\end{tabular}
\end{table}

\begin{figure}[t!]
  \centering
  \includegraphics[width=\linewidth]{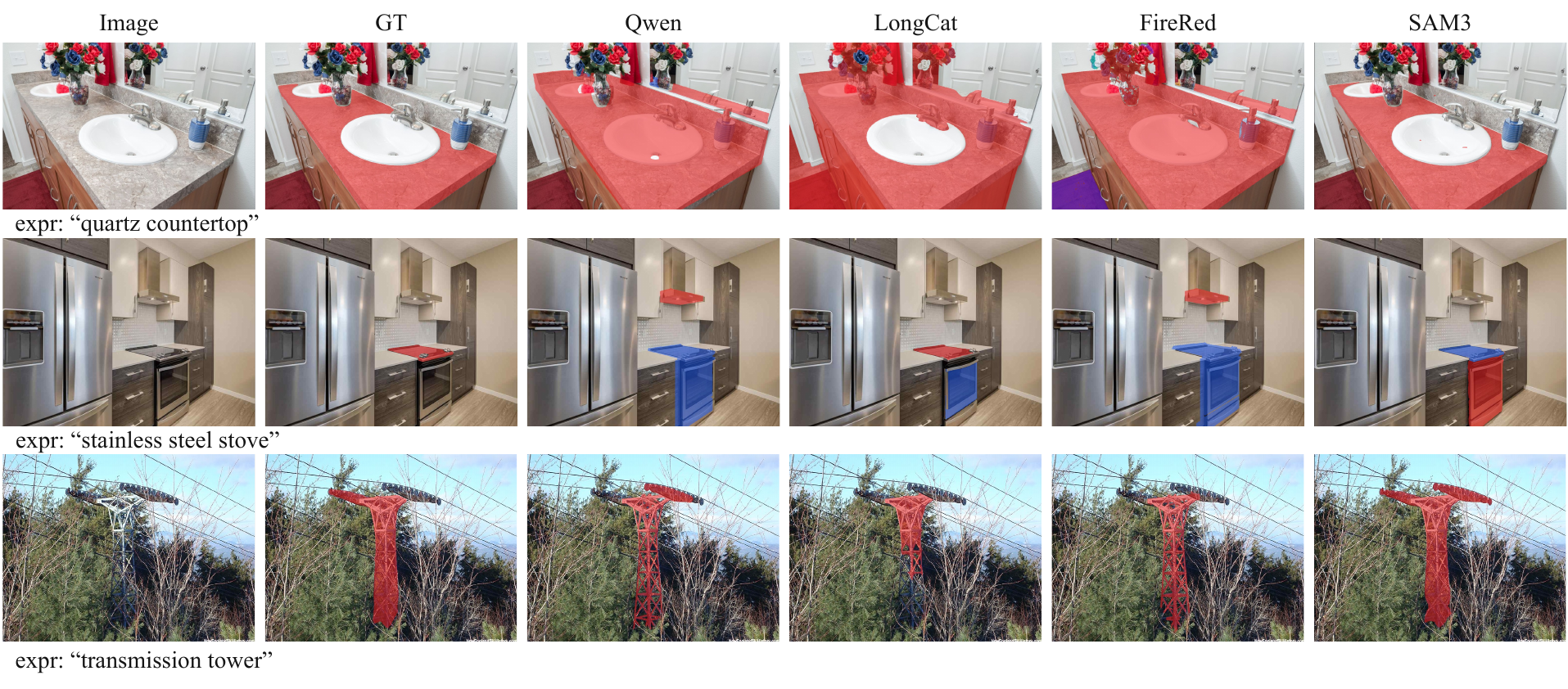}
  \caption{Qualitative open-vocabulary instance segmentation on SA-Co/Gold.}
  \label{fig:vis_saco}
\end{figure}

\textbf{Open-Vocabulary Instance Segmentation (SA-Co).} For each query, we prompt the editor once to repaint the image as a binary mask: all instances matching the phrase are painted pure black, all other regions are painted pure white, and the output should be fully white if the phrase is absent. We take black pixels as foreground and split them into instances using connected components, dropping small components and scoring each instance by area. The model is given no class list or instance count; the query is a single free-form noun phrase that may or may not appear in the image, requiring the editor to perform both concept localization and presence detection. No task-specific head is introduced. We evaluate Qwen-Image-Edit~\citep{qwenimage}, FireRed-Image-Edit~\citep{firered}, and LongCat-Image-Edit~\citep{longcatedit} zero-shot on the attributes split of SA-Co/Gold~\citep{sam3}. We report the three official SA-Co metrics: pmF1, the positive-only micro mask-F1 averaged over IoU thresholds from $0.5$ to $0.95$; IL$\_$MCC, the Matthews correlation coefficient for per-query present/absent prediction; and $\mathrm{cgF1}=100\cdot\mathrm{pmF1}\cdot\mathrm{IL_MCC}$, the headline score combining segmentation quality and presence detection. We report pmF1 and cgF1 as percentages.

\Cref{tab:saco} reports results on the attributes subset. Qwen-Image-Edit performs best among the editors, followed by FireRed and LongCat. While the editors do not yet match dedicated open-vocabulary segmentation models in pmF1, they already produce coherent zero-shot masks for free-form visual concepts without any class list, detector, or segmentation head. The main weakness is open-set presence detection: since the editors tend to paint a mask even when the queried phrase is absent, IL$\_$MCC remains close to zero, which in turn suppresses the headline cgF1. This limitation reflects the absence of an explicit presence classifier rather than a complete failure of RGB-space segmentation; indeed, even Vision~Banana relies on an external MLLM gate to obtain a meaningful IL$\_$MCC. These results suggest that generic image editors can localize named concepts through the RGB editing interface, but reliable absent-case rejection remains an important direction for future improvement. Qualitative visualizations are shown in \Cref{fig:vis_saco}.

\subsection{Generation Tasks: Unifying Inputs as RGB}

\begin{table}[h]
\centering
\caption{Depth-conditioned image generation on the MultiGen-20M depth validation
split. External rows are paper-reported anchors from ControlNet++ and ControlAR;
our editor rows are zero-shot RGB-in/RGB-out editing runs under a shared prompt,
seed~0, \texttt{steps=8}, and \texttt{CFG=2.5}. RMSE-255$\downarrow$ measures
DPT-Large depth consistency, FID$\downarrow$ measures image quality, and
CLIPScore$\uparrow$ measures text alignment.}
\label{tab:depth_generation_main}
\setlength{\tabcolsep}{4pt}
\small
\begin{tabular}{l c c c c}
\toprule
Method & N & RMSE-255$\downarrow$ & FID$\downarrow$ & CLIPScore$\uparrow$ \\
\midrule
\multicolumn{5}{l}{\emph{Trained controllable generation models}}\\
ControlNet++~\citep{controlnetplusplus} & 5000 & 28.32 & 16.66 & 32.09 \\
ControlAR~\citep{controlar} & 5000 & 29.01 & 14.61 & -- \\
ControlNet (SD1.5)~\citep{controlnet} & 5000 & 35.90 & 17.76 & 32.45 \\
GLIGEN~\citep{li2023gligen} & 5000 & 38.83 & 18.36 & 31.75 \\
T2I-Adapter~\citep{t2iadapter} & 5000 & 48.40 & 22.52 & 31.46 \\
Uni-ControlNet~\citep{unicontrolnet} & 5000 & 40.65 & 20.27 & 31.66 \\
UniControl~\citep{unicontrol} & 5000 & 39.18 & 18.66 & 32.45 \\
\midrule
\multicolumn{5}{l}{\emph{Generative image editors (ours, zero-shot, RGB depth input)}}\\
Qwen-Image-Edit & 5000 & \textbf{33.83} & \textbf{19.44} & 30.15 \\
FireRed-Image-Edit& 5000 & 37.55 & 21.43 & \textbf{30.34} \\
LongCat-Image-Edit & 5000 & 36.50 & 27.18 & 28.09 \\
\bottomrule
\end{tabular}
\end{table}

\textbf{Depth-Conditioned Image Generation.} We evaluate depth-to-image generation as a direct test
of RGB-in/RGB-out controllable generation. Following the MultiGen-20M depth
protocol, each validation sample provides an RGB depth map and a text caption. RINO
passes the depth map to an image editor as an ordinary image condition and
instructs it to synthesize the corresponding natural image. No depth encoder,
ControlNet branch, adapter, or task-specific decoder is added; the model only sees
an RGB condition image and a natural-language instruction. For the main
paper-facing comparison, we evaluate Qwen-Image-Edit,
FireRed-Image-Edit, and LongCat-Image-Edit
on the same 5,000-image validation split with the same prompt, seed, and inference
setting: \texttt{steps=8}, \texttt{CFG=2.5}, and a shared short quality prompt. We follow the ControlNet++ evaluation protocol for
depth-conditioned generation~\citep{controlnetplusplus}. Condition fidelity is
measured by extracting depth from generated RGB images with DPT-Large ~\citep{ranftl2021dpt} and
computing RMSE in 0--255 depth space against the input condition. Image quality is
measured by FID against the real validation images, and text alignment by
CLIPScore. Lower RMSE and FID are better; higher CLIPScore is better. The external
rows are paper-reported anchors rather than locally re-scored outputs. We also
note a resolution caveat: the editor outputs in our pipeline are generated at
their native 1024-pixel resolution from 512-pixel MultiGen control maps, whereas
the public depth-control baselines are reported at 512 pixels. A downsample-to-512
sanity check preserves the same qualitative conclusion, but the most directly
aligned cross-method metric is the DPT-Large RMSE-255 condition score.

\begin{figure}[t]
  \centering
  \includegraphics[width=\linewidth]{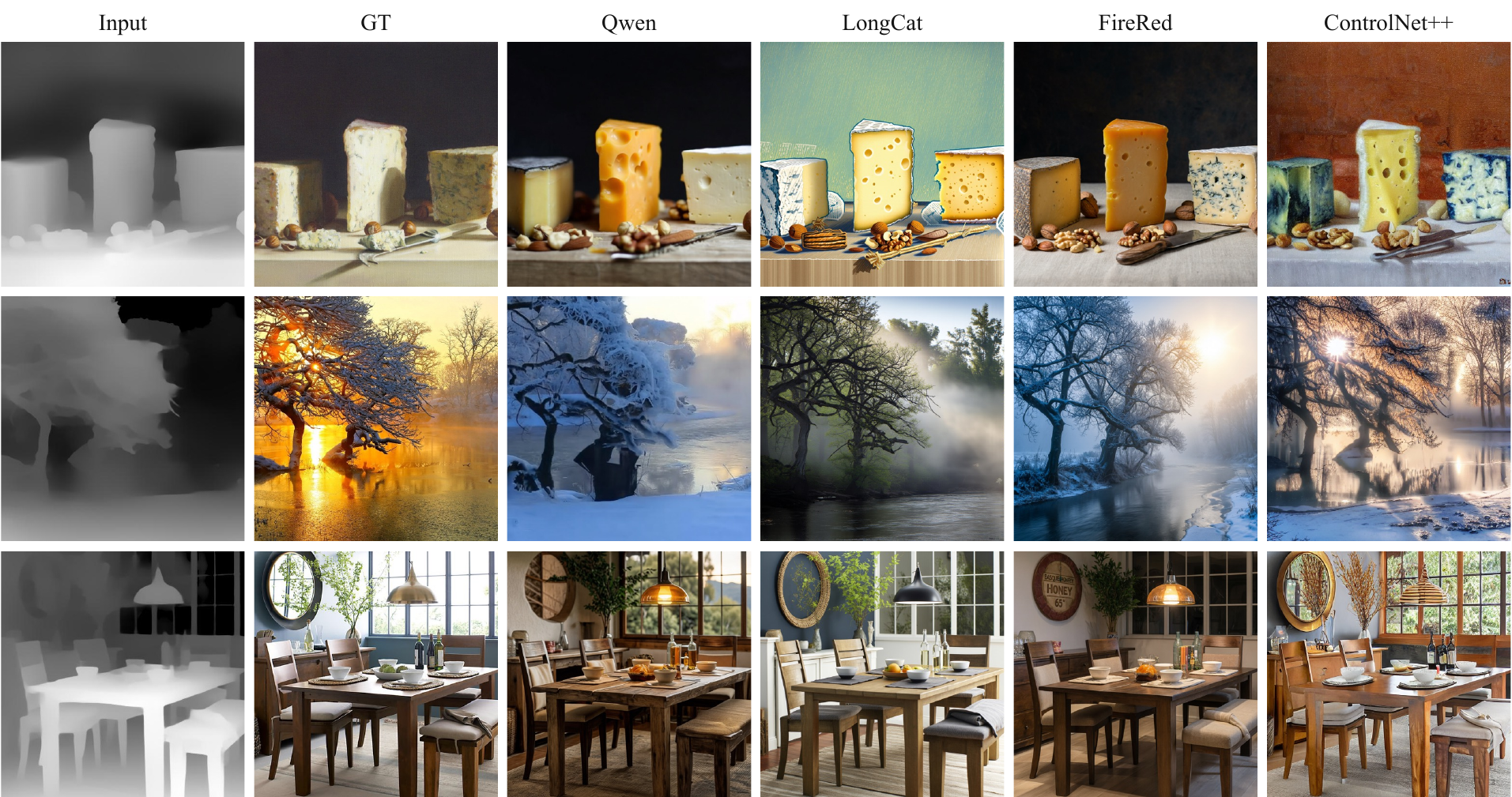}
  \caption{Qualitative depth-conditioned generation on MultiGen-20M. The same RGB
  depth maps and text prompts are given to all three zero-shot editors. The
  three examples show that the RGB-in/RGB-out interface can preserve coarse scene
  geometry, while editor backbones differ in texture realism, semantic fidelity,
  and layout drift.}
  \label{fig:depth_generation_qualitative}
\end{figure}

\Cref{tab:depth_generation_main} compares zero-shot RINO with trained
controllable-generation systems. Among the zero-shot editors, Qwen gives the best
balanced result: RMSE-255 $33.83$, FID $19.44$, and CLIPScore $30.15$. FireRed has
slightly higher CLIPScore ($30.34$) but worse depth fidelity, while LongCat keeps
reasonable RMSE but falls behind in FID and CLIPScore. Qwen remains behind the
strongest trained control models, such as ControlNet++ and ControlAR, which use
dedicated conditioning mechanisms. However, it is competitive with several trained
adapter/control baselines in depth fidelity while using only a zero-shot RGB
editing interface. This supports the central RINO hypothesis: spatial controls can
be consumed as images without introducing a separate encoder or adapter for each
condition type. Qualitative examples in \Cref{fig:depth_generation_qualitative}
show that the editors usually respect coarse depth layout, while differing in
texture realism and semantic faithfulness.

\textbf{Semantic Segmentation Map Conditioned Generation.} Following the ControlNet++ protocol~\citep{controlnetplusplus},
each validation sample provides an RGB semantic segmentation map and a text caption.
RINO treats the segmentation map as an ordinary RGB conditioning image and instructs
the editor to synthesize a natural image whose semantic layout follows the
per-region class colors. No segmentation encoder, ControlNet branch, adapter, or
task-specific decoder is added: the model only sees an RGB condition image and a
natural-language instruction. Because we observe that the richness of the prompt has
a large impact on both image quality and condition fidelity, we re-caption every
validation image once with an open-source VLM and use the resulting caption as the
text prompt for all editors. We evaluate Qwen-Image-Edit~\citep{qwenimage},
FireRed-Image-Edit-1.1~\citep{firered}, and LongCat-Image-Edit~\citep{longcatedit}
on the ADE20K~\citep{ade20k} and COCOStuff~\citep{lin2014microsoft}
validation splits at the native input resolution. Each editor uses its
recommended sampling configuration:
Qwen-Image-Edit with \texttt{steps=40} and \texttt{CFG=4.0},
FireRed-Image-Edit-1.1 with \texttt{steps=40} and \texttt{CFG=4.0},
and LongCat-Image-Edit with \texttt{steps=50} and \texttt{CFG=4.5}.
We follow the ControlNet++ evaluation protocol for
segmentation-conditioned generation~\citep{controlnetplusplus}. Condition fidelity
is measured by re-extracting a semantic segmentation map from each generated image
with a Mask2Former~\citep{mask2former} segmenter and comparing it against the input
condition; we report mean Intersection over Union (mIoU), mean class accuracy
(mAcc), and all-pixel accuracy (aAcc). Image quality is measured by FID against
the real validation images, and text alignment by CLIPScore against the
re-captioned prompt. Lower FID is better; higher mIoU, mAcc, aAcc, and CLIPScore
are better.

\begin{table}[t]
\centering
\caption{Semantic segmentation map conditioned image generation on ADE20K and COCOStuff. ControlNet++ is included as a task-trained
grounded generation reference. Image-editor rows are run by us zero-shot with RGB
segmentation maps and VLM-recaptioned prompts at the native input resolution.}
\label{tab:semantic_seg_map_gen}
\resizebox{\columnwidth}{!}{%
\begin{tabular}{l ccccc ccccc}
\toprule
& \multicolumn{5}{c}{ADE20K} & \multicolumn{5}{c}{COCOStuff} \\
\cmidrule(lr){2-6} \cmidrule(lr){7-11}
Method
& mIoU $\uparrow$
& mAcc $\uparrow$
& aAcc $\uparrow$
& FID $\downarrow$
& CLIP $\uparrow$
& mIoU $\uparrow$
& mAcc $\uparrow$
& aAcc $\uparrow$
& FID $\downarrow$
& CLIP $\uparrow$ \\
\midrule
ControlNet & 32.55 & - & - & 33.28 & 31.53 & 27.46 & - & - & 21.33 & 13.31 \\
ControlNet++ & 43.64 & - & - & 29.49 & 31.96 & 34.56 & - & - & 19.29 & 13.13 \\
\midrule
\multicolumn{11}{l}{\emph{Generative image editors (ours, zero-shot, RGB$\rightarrow$RGB)}} \\
Qwen-Image-Edit &
46.24 & 61.92 & 79.13 & 32.05 & 32.62 &
37.40 & 50.64 & 63.50 & 15.54 & 34.14 \\
FireRed-Image-Edit &
45.82 & 62.82 & 78.14 & 31.70 & 32.67 &
37.55 & 51.20 & 63.25 & 15.94 & 34.07 \\
LongCat-Image-Edit &
46.37 & 61.62 & 78.91 & 36.08 & 32.50 &
36.38 & 48.93 & 61.46 & 20.95 & 33.89 \\
\bottomrule
\end{tabular}%
}
\end{table}

\begin{figure}[h]
  \centering
  \includegraphics[width=\linewidth]{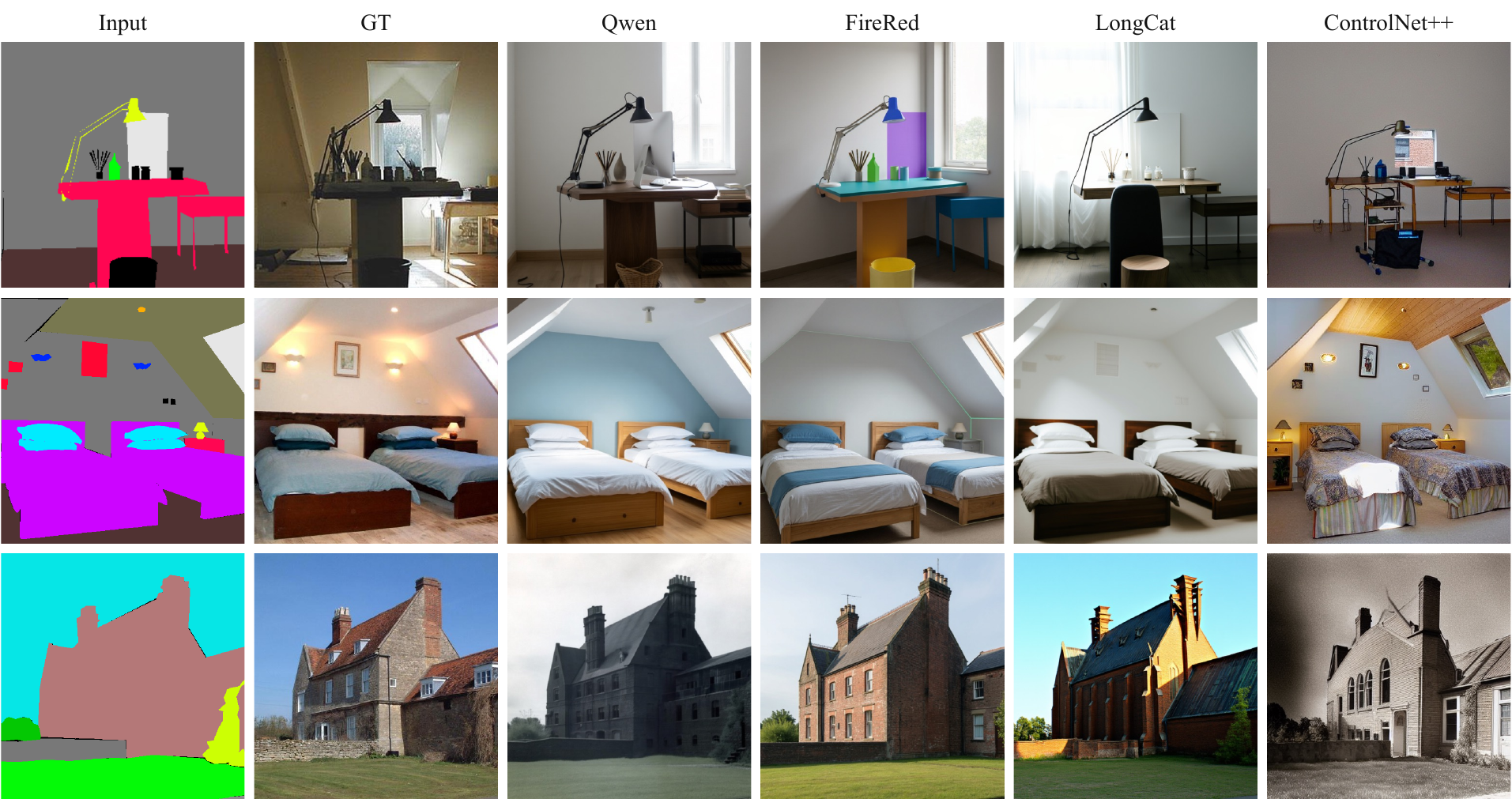}
  \caption{Qualitative semantic segmentation map conditioned generation on ADE20K.
  RINO interface preserves the semantic layout of the input, while
  editor backbones differ in texture realism and small-region fidelity.}
  \label{fig:semseg_generation_qualitative_ade20k}
\end{figure}

\Cref{tab:semantic_seg_map_gen} compares the three zero-shot editors against
ControlNet++ as a task-trained grounded-generation reference on ADE20K and
COCOStuff. All three editors deliver strong condition fidelity: on ADE20K they
cluster tightly around $46$ mIoU (Qwen $46.24$, FireRed $45.82$, LongCat $46.37$),
and on COCOStuff around $37$ mIoU (FireRed $37.55$, Qwen $37.40$, LongCat $36.38$).
Image quality follows the editor backbone rather than the conditioning interface:
Qwen and FireRed are essentially tied on both datasets ($31.70$--$32.05$ FID on
ADE20K, $15.54$--$15.94$ on COCOStuff), while LongCat trails noticeably
($36.08$ and $20.95$ FID). CLIPScore is also tightly clustered across editors,
indicating that the re-captioned prompts are followed comparably well. Overall, a
general-purpose image editor consumes a semantic segmentation map as an ordinary
RGB condition and respects the per-region class layout zero-shot, with image
quality limited by the editor backbone rather than by the lack of a segmentation
adapter. Visualizations are shown in \Cref{fig:semseg_generation_qualitative_ade20k}.

\begin{table}[h]
\centering
\caption{Human-pose conditioned image generation on Human-Art validation. Baselines are quoted from Stable-Pose;
$*$ marks released checkpoints. Our editor rows are zero-shot. \textbf{Bold}: best
baseline per column.}
\label{tab:pose_generation}
\setlength{\tabcolsep}{6pt}
\small
\begin{tabular}{l cc cc c}
\toprule
& \multicolumn{2}{c}{Pose Accuracy} & \multicolumn{2}{c}{Image Quality} & T2I Alignment \\
\cmidrule(lr){2-3}\cmidrule(lr){4-5}\cmidrule(lr){6-6}
Method & CAP$\uparrow$ & PCE$\downarrow$ & FID$\downarrow$ & KID$\downarrow$ & CLIP-score$\uparrow$ \\
\midrule
\multicolumn{6}{l}{\emph{Trained pose-guided generation models}}\\
SD$^{*}$~\citep{stablediffusion}     & 55.71 & 2.30 & 11.53 & 3.36 & \textbf{33.33} \\
T2I-Adapter~\citep{t2iadapter}       & 65.65 & 1.75 & 11.92 & 2.73 & 33.27 \\
ControlNet~\citep{controlnet}        & 69.19 & 1.54 & 11.01 & \textbf{2.23} & 32.65 \\
Uni-ControlNet~\citep{unicontrolnet} & 69.32 & 1.48 & 14.63 & 2.30 & 32.51 \\
GLIGEN~\citep{li2023gligen}          & 69.15 & 1.46 & --    & --   & 32.52 \\
HumanSD~\citep{humansd}              & 69.68 & \textbf{1.37} & \textbf{10.03} & 2.70 & 32.24 \\
Stable-Pose~\citep{stablepose}       & \textbf{70.83} & 1.50 & 11.12 & 2.35 & 32.60 \\
\midrule
\multicolumn{6}{l}{\emph{Generative image editors}}\\
Qwen-Image-Edit        & 65.97 & 1.76 & 36.01 & 5.83 & 32.96 \\
FireRed-Image-Edit     & 68.09 & 1.82 & 10.66 & 5.23 & 34.24 \\
LongCat-Image-Edit     & 55.54 & 2.02 &  7.91 & 6.25 & 32.22 \\
\bottomrule
\end{tabular}
\end{table}

\begin{figure}[t!]
  \centering
  \includegraphics[width=\linewidth]{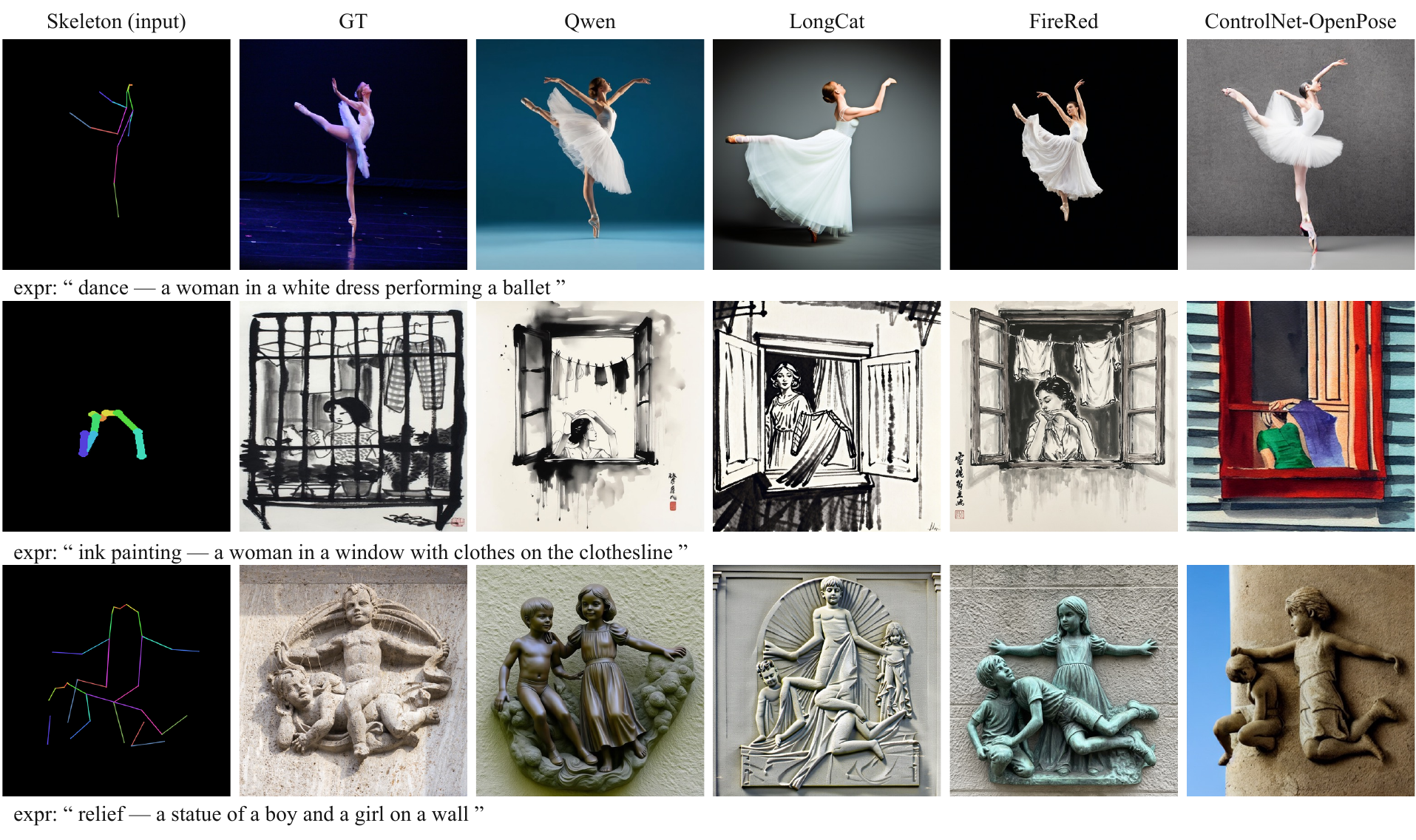}
  \caption{Zero-shot human-pose conditioned generation across styles: input
  skeleton, GT, our editors (Qwen/LongCat/FireRed), and ControlNet-OpenPose.}
  \label{fig:vis_pose_gen}
\end{figure}

\textbf{Human-Pose Conditioned Generation.} We adopt the pose-guided text-to-image protocol of
HumanSD~\citep{humansd} and Stable-Pose~\citep{stablepose} on the Human-Art ~\citep{ju2023humanart}
validation split. For each sample, the
ground-truth COCO-17 keypoints are rendered into an OpenPose-style colored skeleton ($16$ limb segments on a black canvas).  The editor receives only this RGB skeleton and the associated caption, and is instructed to align each body part with the corresponding colored bone while omitting the skeleton overlay from the synthesized image. No pose encoder, ControlNet branch, or trainable adapter is introduced. We follow the HumanSD/Stable-Pose evaluation protocol. Pose fidelity is assessed by re-estimating the pose of each generated image with HigherHRNet ~\citep{cheng2020higherhrnet} and comparing it to the conditioning skeleton through
COCOevalSimilarity, yielding a
cosine-similarity-based average precision (CAP), and a people-count error (PCE). Image quality is measured by FID and KID against the real Human-Art images, and text alignment by CLIP-Score. Editor outputs are scored with the public evaluation code under which the baseline numbers were reported.

\Cref{tab:pose_generation} and \Cref{fig:vis_pose_gen} show that RINO achieves highly competitive zero-shot performance for human-pose conditioned image generation. Although our editors are not trained on this task, their results are already comparable to in-domain pose-guided generation models, demonstrating that generic image editors can effectively interpret RGB skeleton inputs as spatial generation conditions. Qualitatively, RINO often produces images that better follow the input pose than ControlNet-based baselines, especially in cases involving unusual body configurations or diverse artistic styles. The generated images also maintain strong text-image alignment and visual realism, suggesting that the RGB-in/RGB-out interface provides an effective and flexible alternative to task-specific pose encoders for conditioned generation.

\begin{table}[t]
\centering
\caption{Instance map conditioned image generation on COCO2017 validation.
InstanceDiffusion is included as a task-trained grounded generation reference.
Image-editor rows are run by us zero-shot with RGB instance maps and
VLM-recaptioned prompts at the native input resolution.}
\label{tab:instance_map_gen}
\setlength{\tabcolsep}{3pt}
\small
\begin{tabular}{@{}lcccc@{}}
\toprule
Method & AP$\uparrow$ & AP$_{50}\uparrow$ & Average Recall$\uparrow$ & FID$\downarrow$ \\
\midrule
\multicolumn{5}{l}{\emph{Reported task-trained grounded generation result}}\\
InstanceDiffusion~\citep{wang2024instancediffusion} & 27.1 & 50.0 & 38.1 & 25.5 \\
\midrule
\multicolumn{5}{l}{\emph{Generative image editors (ours, zero-shot, RGB instance map input)}}\\
Qwen-Image-Edit~\citep{qwenimage} & 25.4 & 44.3 & 40.8 & 15.3 \\
FireRed-Image-Edit~\citep{firered} & 23.2 & 42.3 & 37.4 & 16.0 \\
LongCat-Image-Edit~\citep{longcatedit} & 28.0 & 46.3 & 44.2 & 18.5 \\
\bottomrule
\end{tabular}
\end{table}

\begin{figure}[t!]
  \centering
  \includegraphics[width=\linewidth]{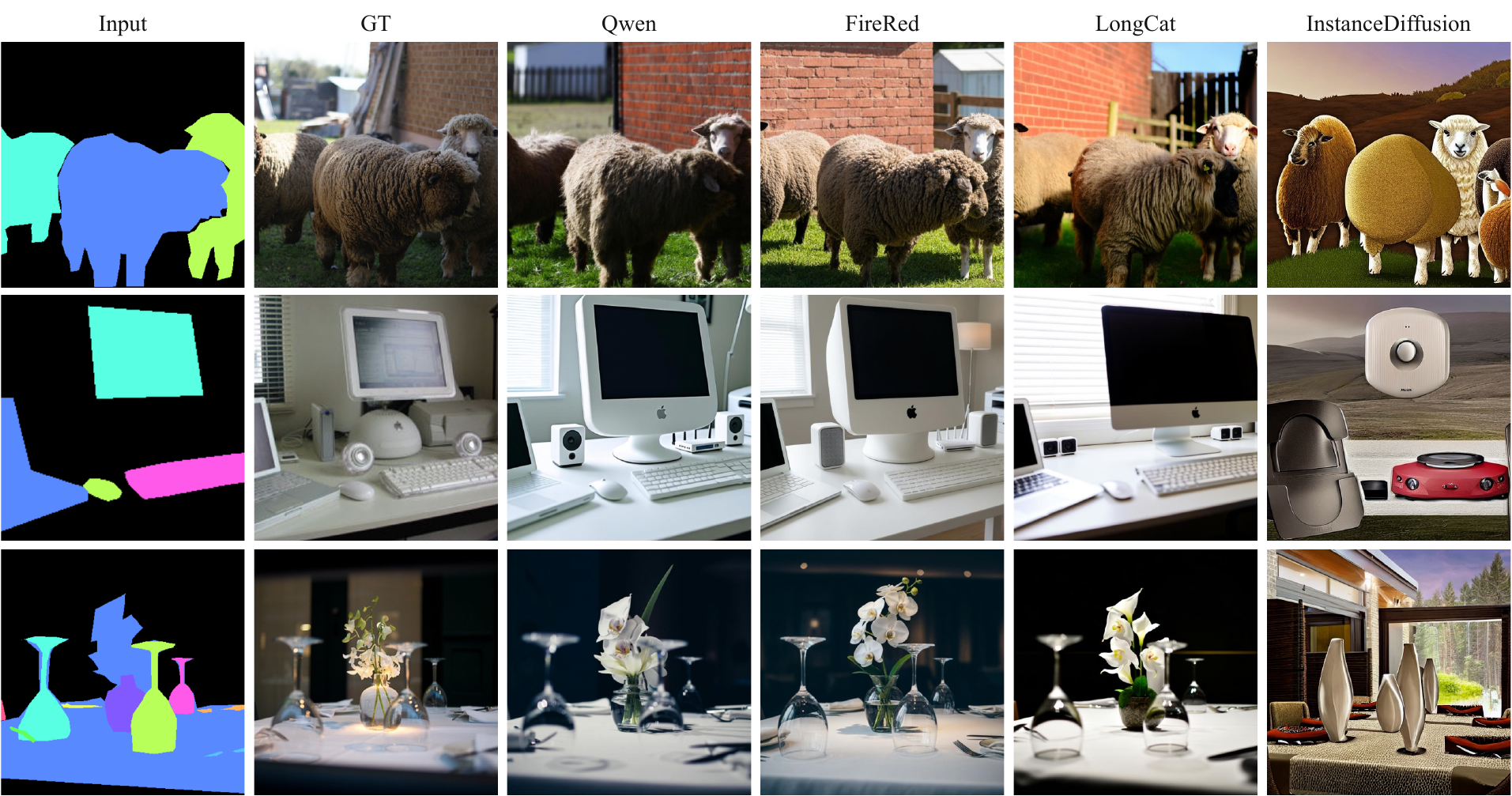}
  \caption{Qualitative instance map conditioned generation on COCO2017
  validation. The same RGB instance maps and re-captioned prompts are given to
  all three zero-shot editors. The examples show that the RGB-in/RGB-out
  interface preserves per-instance placements without an instance encoder,
  while editor backbones differ in texture realism and small-instance fidelity.}
  \label{fig:instance_generation_qualitative}
\end{figure}

\textbf{Instance Map Conditioned Generation.} Following the InstanceDiffusion
protocol~\citep{wang2024instancediffusion}, each validation sample provides a
per-instance map together with a text caption. Unlike a semantic segmentation
map, this condition distinguishes individual object instances of the same class
rather than just per-category regions. RINO renders the instance map as an
ordinary RGB conditioning image and instructs the editor to synthesize a natural
image whose object instances follow the input layout. No instance encoder,
grounding module, detector head, or trainable adapter is introduced. As in the
segmentation-map setting, we re-caption every validation image once with an
open-source VLM and use that caption as the prompt for all editors, because we
find caption richness materially affects both image quality and the
recognizability of generated instances. We evaluate
Qwen-Image-Edit, FireRed-Image-Edit,
and LongCat-Image-Edit on the COCO2017~\citep{wang2024instancediffusion}
validation split at the native input resolution. Each editor uses its
recommended sampling configuration:
Qwen-Image-Edit with \texttt{steps=40} and \texttt{CFG=4.0},
FireRed-Image-Editwith \texttt{steps=40} and \texttt{CFG=4.0},
and LongCat-Image-Edit with \texttt{steps=50} and \texttt{CFG=4.5}. We follow the InstanceDiffusion evaluation
protocol~\citep{wang2024instancediffusion}. Condition fidelity is measured by
running a pretrained YOLOv8m-Seg~\citep{yolov8} detector on the generated images and comparing
its detected instance masks against the input condition with COCO's official
evaluation metrics. We report mask AP (IoU $0.5{:}0.05{:}0.95$),
AP$_{50}$ (IoU\,$\ge0.5$), and Average Recall (AR), which together measure both
the localization precision and the recall of the requested instances. Image
quality is measured by FID against the real COCO2017 validation images.
Higher AP, AP$_{50}$, and AR are better; lower FID is better.

Table~\ref{tab:instance_map_gen} compares the three zero-shot editors with
InstanceDiffusion as a task-trained grounded-generation reference. The editors
follow the per-instance layout closely: LongCat reaches the highest AP ($28.0$)
and AR ($44.2$), surpassing the trained InstanceDiffusion reference on both
metrics, while Qwen ($25.4$ AP, $44.3$ AP$_{50}$, $40.8$ AR) and FireRed
($23.2$ AP, $42.3$ AP$_{50}$, $37.4$ AR) remain within a few points of it. At
the stricter AP$_{50}$ threshold the trained reference still leads ($50.0$ vs.\
$42$--$46$ for the editors), indicating that precise instance boundaries are
harder to recover than coarse instance placement. Image quality is consistently
better than the trained reference across all editors (FID $15.3$--$18.5$ vs.\
$25.5$), reflecting the strength of the modern editor backbones. Overall, a
general-purpose image editor can consume a per-instance RGB layout zero-shot
and produce images that respect individual instance placements while
maintaining higher visual quality than the task-trained baseline. Visualizations
are shown in \Cref{fig:instance_generation_qualitative}.

\begin{table}[htbp!]
\centering
\caption{Canny-conditioned generation on MultiGen-20M validation (5000 images,
$512^2$, 4 samples each). Edge controllability is F1 (\texttt{cv2.Canny(100,200)},
$\times100$); image quality is FID; text alignment is
CLIP-Score (CLIP ViT-B/16) --- all following the ControlNet++ protocol, with FID
and CLIP averaged over the 4 generated groups. Our \textsc{Qwen-Image-Edit} rows
are \emph{zero-shot}; trained rows are paper-reported anchors from
ControlNet++~\citep{controlnetplusplus}. \textbf{Bold}: best zero-shot per column.}
\label{tab:canny-controllability}
\begin{tabular}{lccc}
\toprule
Method & F1 $\uparrow$ & FID $\downarrow$ & CLIPScore $\uparrow$ \\
\midrule
\multicolumn{4}{l}{\emph{Zero-shot (ours and baselines)}} \\
Raw Caption-only                          & 6.75  & 71.61 & \textbf{27.78} \\
Raw Caption + Canny Caption               & 7.79  & 59.39 & 23.98 \\
\textsc{Qwen-Image-Edit} (canny + caption) & \textbf{14.90} & \textbf{58.30} & 27.31 \\
\midrule
\multicolumn{4}{l}{\emph{Trained for canny conditioning}~\citep{controlnetplusplus}} \\
T2I-Adapter~\citep{t2iadapter}            & 23.65 & 15.96 & 31.71 \\
GLIGEN~\citep{li2023gligen}               & 26.94 & 18.89 & 31.77 \\
Uni-ControlNet~\citep{unicontrolnet}      & 27.32 & 17.14 & 31.84 \\
UniControl~\citep{unicontrol}             & 30.82 & 19.94 & 31.97 \\
ControlNet (SD1.5)~\citep{controlnet}     & 34.65 & 14.73 & 32.15 \\
ControlNet++ (SD1.5)~\citep{controlnetplusplus} & 37.04 & 18.23 & 31.87 \\
\bottomrule
\end{tabular}
\end{table}

\textbf{Canny-Conditioned Image Generation.} Following the MultiGen-20M protocol~\citep{controlnetplusplus}, each sample consists of a Canny edge map (white edges on a black background) and a
text caption. RINO treats the edge map as an ordinary RGB conditioning image and
instructs the editor to synthesize a natural image whose object boundaries and
spatial layout conform to the edges, adding no edge encoder, ControlNet branch,
adapter, or task-specific decoder. We evaluate Qwen-Image-Edit~\citep{qwenimage}
on the $5{,}000$-image validation split at $512{\times}512$, drawing $4$ samples
per condition with \texttt{steps=30} and \texttt{CFG=4.0} under a
shared instruction prompt. All reported numbers are means over the $4$ samples per image; no best-of-$N$ selection is used. We follow the ControlNet++ protocol for the
Canny condition~\citep{controlnetplusplus}. Condition fidelity is measured by
re-extracting a Canny edge map from each generated image and comparing it to the input condition;
because Canny is a hard (binary) edge, the controllability metric is the per-pixel
edge F1-Score, extracted with OpenCV
\texttt{cv2.Canny(100,200)} to match the protocol exactly. We
additionally report image quality as FID against the real
validation images and text alignment as CLIP-Score (using CLIP ViT-B/16), both averaged
over the $4$ generated groups and computed with the same implementations as
ControlNet++; the re-extracted edge maps are excluded from the FID image set.

\Cref{tab:canny-controllability} reports all three metrics. On
controllability, zero-shot Qwen-Image-Edit reaches a Canny F1 of $14.90$, about $0.40\times$ the purpose-trained
ControlNet++ ($37.04$) and below every trained prior ($23$--$37$). The same gap
holds on image quality and text alignment: its FID of $59.4$ and CLIP-Score
of $27.3$ trail the trained controllers (FID $15$--$20$, CLIP $\approx32$).
Zero-shot canny generation therefore trails purpose-trained controllers on all
three axes --- edge controllability, distributional realism, and prompt alignment
--- as expected for a general instruction editor with no edge-conditioned training
and no dedicated conditioning branch. The result nonetheless supports the central
RINO hypothesis in the harder binary-edge regime: a spatial control can be
consumed as an ordinary image, with above-chance controllability, without any
task-specific adapter.

\begin{figure}[t]
  \centering
  \includegraphics[width=\linewidth]{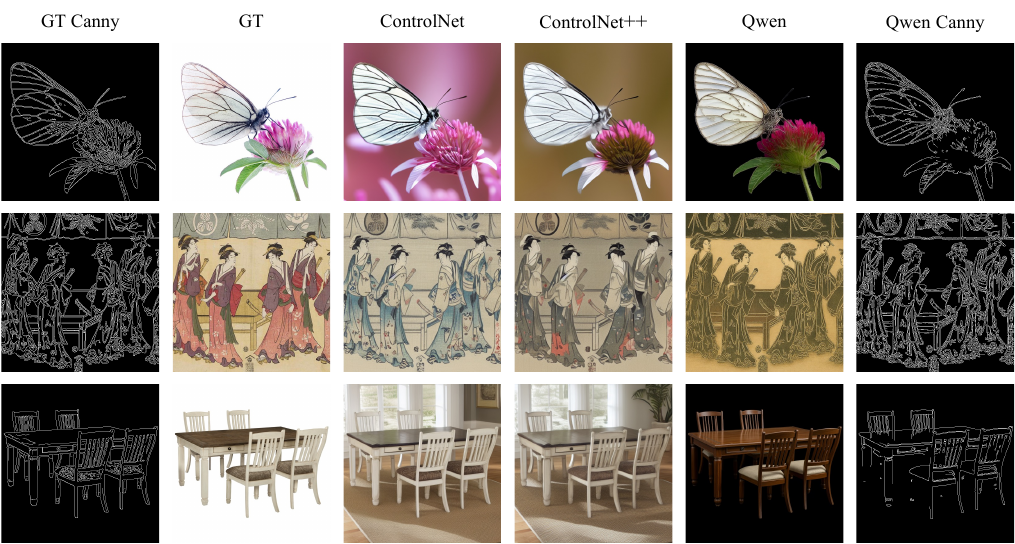}
  \caption{Qualitative canny-conditioned generation on  MultiGen-20M.}
  \label{fig:canny_qualitative}
\end{figure}

To verify that this controllability comes from the Canny \emph{image} rather than
the caption, we ablate the condition while holding the editor fixed (top block of
\Cref{tab:canny-controllability}). Removing the edge map and keeping only the
caption (\emph{Raw Caption-only}) collapses fidelity to $6.75$, and first
\emph{describing} the edge map in words with a vision-language model
(Qwen2.5-VL-7B) and feeding that text instead of the image (\emph{Raw Caption +
Canny Caption}) reaches only $7.79$ --- both far below the $14.90$ from the edge image itself. This implies that the Canny image carries
pixel-level spatial information that a language bottleneck cannot substitute. The
quality metrics add nuance: supplying structure (the Canny image or its VLM
description) improves FID over caption-only ($71.6\to58$--$59$), i.e.\ a spatial
prior yields more realistic images, but routing it through a long VLM description
lowers CLIP-Score ($27.8\to24.0$), as the appended layout text competes with the
caption for prompt alignment. Visualization results are shown in \Cref{fig:canny_qualitative}.

\section{Related Work}
\textbf{Visual estimation models.}
Estimation tasks map a natural image to a structured signal, and each task we study has a mature line of specialists that defines its performance ceiling andserves as our reference. Monocular depth estimation progresses from mixing datasets for scale- and shift-invariant relative depth~\citep{ranftl2020midas,midasv31,zoedepth} to large-scale data-driven and metric-aware systems~\citep{depthanything,depthanythingv2,depthanythingv3,moge2,unik3d}, and pretrained diffusion generators are themselves fine-tuned into strong depth and normal estimators~\citep{ke2024marigold,ye2024stablenormal,lotus2,DSINE}. Segmentation converges on the mask-classification view, where a transformer predicts a set of labeled masks to cover semantic, instance, and panoptic segmentation at
once~\citep{cheng2021per,mask2former,oneformer,eomt,vitp,maskdino,xie2021segformer},
with an open-vocabulary branch that names novel categories via language embeddings~\citep{xu2022groupvit,dong2023maskclip,yu2023convolutions,barsellotti2024training,xu2023learning}.
The remaining tasks follow the same pattern of dedicated architectures: pose estimation~\citep{vitpose,yolopose,sam3dbody, yang2026fastsam3dbody}, detection and instance segmentation with text-grounded open-set
variants~\citep{maskrcnn,cascadercnn,maskdino,groundingdino}, and referring segmentation~\citep{lavt,uninext,oneref}, many built on general-purpose backbones~\citep{clip,dinov2,sam} but still read out through a task-specific head. Unification here stays \emph{within} a family, with each keeping its own encoder, decoder, and loss.

\textbf{Controllable image generation.}
Generation tasks move in the opposite direction, synthesizing a natural image from a structured condition---Canny edges, depth, semantic or instance maps, boxes, or poses. ControlNet~\citep{controlnet} conditions a text-to-image
backbone by attaching a trainable encoder copy that injects the control signal, and later work improves fidelity or extends it to autoregressive backbones~\citep{controlnetplusplus,controlar}; adapter methods instead learn small per-condition side networks~\citep{t2iadapter}, and unified frameworks route
many condition types through a shared module~\citep{unicontrolnet,unicontrol}.
Layout- and instance-grounded generation~\citep{li2023gligen,wang2024instancediffusion,suleyman2025grounding} and pose-guided synthesis~\citep{stablepose,humansd} follow the same recipe. The recurring ingredient is a per-condition encoder or adapter trained to inject each new signal. Estimation and generation are thus near-mirror images---one reads a structured signal out of an image, the other writes one in---yet the two traditionally use separate architectures and output spaces and are never unified under a shared interface.

\textbf{Toward a unified model for estimation and generation.}
Vision Banana~\citep{visionbanana} is the first to bridge the two directions: parameterizing the output space of vision tasks as RGB images reframes perception as image generation, and instruction-tuning a single generator makes it
competitive with zero-shot specialists such as Segment Anything~\citep{sam3} and
the Depth Anything series~\citep{depthanything,depthanythingv2,depthanythingv3}, with open-source editors shown to exhibit similar behavior~\citep{tencentbanana}. This casts RGB
as a candidate common interface for both reading and writing images, analogous to text in language models. Vision Banana, however, is built on a proprietary backbone, leaving open how far this interface carries over to open, publicly
released models. RINO shares its RGB-as-interface premise and pushes it to this setting: we take open-source image editors as a frozen black box, add only parameter-free RGB conversions, and place estimation and conditioned generation under one RGB-to-RGB formulation across 25 tasks, measuring how closely this fully zero-shot system approaches the specialists above.

\section{Conclusion}

RINO shows that RGB can act as a common interface for both perception and generation: expressing all visual inputs and outputs as RGB lets a single frozen image editor handle a broad range of tasks without any task-specific encoders, decoders, adapters, or auxiliary parameters. Fully zero-shot and without any training, RINO-Zero approaches in-domain specialist models across 25 tasks spanning dense estimation, 3D geometry, and conditioned generation, and offers a zero-shot alternative to adapter-based controllable generation. Its accuracy remains bounded by the visual priors of the editing backbone and by the parameter-free RGB read-out, leaving a gap to specialists on some tasks; lightly instruction-tuning editors on RGB-formatted visual signals, so that the RINO interface is present during training, is a natural way to close it. These findings point to RGB as a shared interface for vision and invite its extension to 3D vision, video, and world models.

\bibliographystyle{plainnat}
\bibliography{references}

@article{visionbanana,
  title={Image Generators are Generalist Vision Learners},
  author={Valentin Gabeur and Shangbang Long and Songyou Peng and Paul Voigtlaender and Shuyang Sun and Yanan Bao and Karen Truong and Zhicheng Wang and Wenlei Zhou and Jonathan T. Barron and Kyle Genova and Nithish Kannen and Sherry Ben and Yandong Li and Mandy Guo and Suhas Yogin and Yiming Gu and Huizhong Chen and Oliver Wang and Saining Xie and Howard Zhou and Kaiming He and Thomas Funkhouser and Jean-Baptiste Alayrac and Radu Soricut},
  journal={arXiv preprint arXiv:2604.20329},
  year={2026}
}

@article{tencentbanana,
  title={Open-Source Image Editing Models Are Zero-Shot Vision Learners},
  author={Wei Liu and Jiaxin Lin and Rui Chen},
  journal={arXiv preprint arXiv:2605.04566},
  year={2026}
}

@article{qwenimage,
  title={Qwen-Image Technical Report},
  author={Chenfei Wu and Jiahao Li and Jingren Zhou and Junyang Lin and Kaiyuan Gao and Kun Yan and Sheng-ming Yin and Shuai Bai and Xiao Xu and Yilei Chen and Yuxiang Chen and Zecheng Tang and Zekai Zhang and Zhengyi Wang and An Yang and Bowen Yu and Chen Cheng and Dayiheng Liu and Deqing Li and Hang Zhang and Hao Meng and Hu Wei and Jingyuan Ni and Kai Chen and Kuan Cao and Liang Peng and Lin Qu and Minggang Wu and Peng Wang and Shuting Yu and Tingkun Wen and Wensen Feng and Xiaoxiao Xu and Yi Wang and Yichang Zhang and Yongqiang Zhu and Yujia Wu and Yuxuan Cai and Zenan Liu},
  journal={arXiv preprint arXiv:2508.02324},
  year={2025}
}

@inproceedings{controlnet,
  title={Adding Conditional Control to Text-to-Image Diffusion Models},
  author={Lvmin Zhang and Anyi Rao and Maneesh Agrawala},
  booktitle={ICCV},
  year={2023}
}

@inproceedings{sam,
  title={Segment Anything},
  author={Alexander Kirillov and Eric Mintun and Nikhila Ravi and Hanzi Mao and Chloe Rolland and Laura Gustafson and Tete Xiao and Spencer Whitehead and Alexander C. Berg and Wan-Yen Lo and Piotr Doll{\'a}r and Ross Girshick},
  booktitle={ICCV},
  year={2023}
}

@inproceedings{depthanything,
  title={Depth Anything: Unleashing the Power of Large-Scale Unlabeled Data},
  author={Lihe Yang and Bingyi Kang and Zilong Huang and Xiaogang Xu and Jiashi Feng and Hengshuang Zhao},
  booktitle={CVPR},
  year={2024}
}

@inproceedings{stablediffusion,
  title={High-Resolution Image Synthesis with Latent Diffusion Models},
  author={Robin Rombach and Andreas Blattmann and Dominik Lorenz and Patrick Esser and Bj{\"o}rn Ommer},
  booktitle={CVPR},
  year={2022}
}

@inproceedings{clip,
  title={Learning Transferable Visual Models From Natural Language Supervision},
  author={Alec Radford and Jong Wook Kim and Chris Hallacy and Aditya Ramesh and Gabriel Goh and Sandhini Agarwal and Girish Sastry and Amanda Askell and Pamela Mishkin and Jack Clark and Gretchen Krueger and Ilya Sutskever},
  booktitle={ICML},
  year={2021}
}

@article{dinov2,
  title={DINOv2: Learning Robust Visual Features without Supervision},
  author={Maxime Oquab and Timoth{\'e}e Darcet and Th{\'e}o Moutakanni and Huy Vo and Marc Szafraniec and Vasil Khalidov and Pierre Fernandez and Daniel Haziza and Francisco Massa and Alaaeldin El-Nouby and Mahmoud Assran and Nicolas Ballas and Wojciech Galuba and Russell Howes and Po-Yao Huang and Shang-Wen Li and Ishan Misra and Michael Rabbat and Vasu Sharma and Gabriel Synnaeve and Hu Xu and Herv{\'e} J{\'e}gou and Julien Mairal and Patrick Labatut and Armand Joulin and Piotr Bojanowski},
  journal={Transactions on Machine Learning Research},
  year={2024}
}

@inproceedings{ade20k,
  title={Scene Parsing through ADE20K Dataset},
  author={Bolei Zhou and Hang Zhao and Xavier Puig and Sanja Fidler and Adela Barriuso and Antonio Torralba},
  booktitle={CVPR},
  year={2017}
}

@article{ranftl2020midas,
  title={Towards Robust Monocular Depth Estimation: Mixing Datasets for Zero-Shot Cross-Dataset Transfer},
  author={Ren{\'e} Ranftl and Katrin Lasinger and David Hafner and Konrad Schindler and Vladlen Koltun},
  journal={IEEE Transactions on Pattern Analysis and Machine Intelligence (TPAMI)}, year={2022}}

@inproceedings{ke2024marigold,
  title={Repurposing Diffusion-Based Image Generators for Monocular Depth Estimation},
  author={Bingxin Ke and Anton Obukhov and Shengyu Huang and Nando Metzger and Rodrigo Caye Daudt and Konrad Schindler},
  booktitle={CVPR}, year={2024}}

@article{longcatedit,
  title={LongCat-Image Technical Report},
  author={Hanghang Ma and Haoxian Tan and Jiale Huang and Junqiang Wu and Jun-Yan He and Lishuai Gao and Songlin Xiao and Xiaoming Wei and Xiaoqi Ma and Xunliang Cai and Yayong Guan and Jie Hu},
  journal={arXiv preprint arXiv:2512.07584}, year={2025}}

@article{zoedepth,
  title={ZoeDepth: Zero-shot Transfer by Combining Relative and Metric Depth},
  author={Shariq Farooq Bhat and Reiner Birkl and Diana Wofk and Peter Wonka and Matthias M{\"u}ller},
  journal={arXiv preprint arXiv:2302.12288}, year={2023}}

@article{moge2,
  title={MoGe-2: Accurate Monocular Geometry with Metric Scale and Sharp Details},
  author={Ruicheng Wang and Sicheng Xu and Yue Dong and Yu Deng and Jianfeng Xiang and Zelong Lv and Guangzhong Sun and Xin Tong and Jiaolong Yang},
  journal={arXiv preprint arXiv:2507.02546}, year={2025}}

@inproceedings{unik3d,
  title={UniK3D: Universal Camera Monocular 3D Estimation},
  author={Luigi Piccinelli and Christos Sakaridis and Mattia Segu and Yung-Hsu Yang and Siyuan Li and Wim Abbeloos and Luc Van Gool},
  booktitle={CVPR}, year={2025}}

@article{depthanythingv3,
  title={Depth Anything 3: Recovering the Visual Space from Any Views},
  author={Haotong Lin and Sili Chen and Jun Hao Liew and Donny Y. Chen and Zhenyu Li and Guang Shi and Jiashi Feng and Bingyi Kang},
  journal={arXiv preprint arXiv:2511.10647}, year={2025}}

@article{midasv31,
  title={MiDaS v3.1 -- A Model Zoo for Robust Monocular Relative Depth Estimation},
  author={Birkl, Reiner and Wofk, Diana and M{\"u}ller, Matthias},
  journal={arXiv preprint arXiv:2307.14460},
  year={2023}
}

@article{depthanythingv2,
  title={Depth Anything V2},
  author={Yang, Lihe and Kang, Bingyi and Huang, Zilong and Zhao, Zhen and Xu, Xiaogang and Feng, Jiashi and Zhao, Hengshuang},
  journal={Advances in Neural Information Processing Systems (NeurIPS)},
  volume={37},
  pages={21875--21911},
  year={2024}
}

@inproceedings{DSINE,
  title={Rethinking Inductive Biases for Surface Normal Estimation},
  author={Bae, Gwangbin and Davison, Andrew J.},
  booktitle={Proceedings of the IEEE/CVF Conference on Computer Vision and Pattern Recognition (CVPR)},
  pages={9535--9545},
  year={2024}
}

@article{ye2024stablenormal,
  title={StableNormal: Reducing Diffusion Variance for Stable and Sharp Normal},
  author={Ye, Chongjie and Qiu, Lingteng and Gu, Xiaodong and Zuo, Qi and Wu, Yushuang and Dong, Zilong and Bo, Liefeng and Xiu, Yuliang and Han, Xiaoguang},
  journal={ACM Transactions on Graphics (TOG)},
  volume={43},
  number={6},
  pages={1--18},
  year={2024}
}

@article{lotus2,
  title={Lotus-2: Advancing Geometric Dense Prediction with Powerful Image Generative Model},
  author={He, Jing and Li, Haodong and Sheng, Mingzhi and Chen, Ying-Cong},
  journal={arXiv preprint arXiv:2512.01030},
  year={2025}
}

@misc{firered,
  title={{FireRed-Image-Edit-1.0}},
  author={{FireRed Team}},
  howpublished={Hugging Face model, \texttt{FireRedTeam/FireRed-Image-Edit-1.0}},
  year={2025}
}

@inproceedings{mask2former,
  title={Masked-attention Mask Transformer for Universal Image Segmentation},
  author={Bowen Cheng and Ishan Misra and Alexander G. Schwing and Alexander Kirillov and Rohit Girdhar},
  booktitle={CVPR},
  year={2022}
}

@inproceedings{oneformer,
  title={OneFormer: One Transformer to Rule Universal Image Segmentation},
  author={Jitesh Jain and Jiachen Li and Mang Tik Chiu and Ali Hassani and Nikita Orlov and Humphrey Shi},
  booktitle={CVPR},
  year={2023}
}

@inproceedings{eomt,
  title={Your ViT is Secretly an Image Segmentation Model},
  author={Tommie Kerssies and Niccol{\`o} Cavagnero and Alexander Hermans and Narges Norouzi and Giuseppe Averta and Bastian Leibe and Gijs Dubbelman and Daan de Geus},
  booktitle={CVPR},
  year={2025}
}

@article{vitp,
  title={The Missing Point in Vision Transformers for Universal Image Segmentation},
  author={Sajjad Shahabodini and Mobina Mansoori and Farnoush Bayatmakou and Jamshid Abouei and Konstantinos N. Plataniotis and Arash Mohammadi},
  journal={arXiv preprint arXiv:2505.19795},
  year={2025}
}

@misc{yang2026fastsam3dbody,
      title={Fast SAM 3D Body: Accelerating SAM 3D Body for Real-Time Full-Body Human Mesh Recovery}, 
      author={Timing Yang and Sicheng He and Hongyi Jing and Jiawei Yang and Zhijian Liu and Chuhang Zou and Yue Wang},
      year={2026},
      eprint={2603.15603},
      archivePrefix={arXiv},
      primaryClass={cs.CV},
      url={https://arxiv.org/abs/2603.15603}, 
}

@inproceedings{vitpose,
  title={ViTPose: Simple Vision Transformer Baselines for Human Pose Estimation},
  author={Yufei Xu and Jing Zhang and Qiming Zhang and Dacheng Tao},
  booktitle={NeurIPS},
  year={2022}
}

@inproceedings{maskrcnn,
  title={Mask R-CNN},
  author={Kaiming He and Georgia Gkioxari and Piotr Doll{\'a}r and Ross Girshick},
  booktitle={ICCV},
  year={2017}
}

@inproceedings{yolopose,
  title={YOLO-Pose: Enhancing YOLO for Multi Person Pose Estimation Using Object Keypoint Similarity Loss},
  author={Debapriya Maji and Soyeb Nagori and Manu Mathew and Deepak Poddar},
  booktitle={CVPR Workshops (CVPRW)},
  year={2022}
}

@article{sam3dbody,
  title={SAM 3D Body: Robust Full-Body Human Mesh Recovery},
  author={Xitong Yang and Devansh Kukreja and Don Pinkus and Anushka Sagar and Taosha Fan and Jinhyung Park and Soyong Shin and Jinkun Cao and Jiawei Liu and Nicolas Ugrinovic and Matt Feiszli and Jitendra Malik and Piotr Doll{\'a}r and Kris Kitani},
  journal={arXiv preprint arXiv:2602.15989},
  year={2026}
}

@inproceedings{maskdino,
  title={Mask {DINO}: Towards A Unified Transformer-based Framework for Object Detection and Segmentation},
  author={Feng Li and Hao Zhang and Huaizhe Xu and Shilong Liu and Lei Zhang and Lionel M. Ni and Heung-Yeung Shum},
  booktitle={CVPR},
  year={2023}
}

@article{cascadercnn,
  title={Cascade R-CNN: High Quality Object Detection and Instance Segmentation},
  author={Zhaowei Cai and Nuno Vasconcelos},
  journal={IEEE Transactions on Pattern Analysis and Machine Intelligence (TPAMI)},
  year={2021}
}

@inproceedings{lavt,
  title={Lavt: Language-aware vision transformer for referring image segmentation},
  author={Yang, Zhao and Wang, Jiaqi and Tang, Yansong and Chen, Kai and Zhao, Hengshuang and Torr, Philip HS},
  booktitle={Proceedings of the IEEE/CVF conference on computer vision and pattern recognition},
  pages={18155--18165},
  year={2022}
}

@inproceedings{uninext,
  title={Universal Instance Perception as Object Discovery and Retrieval},
  author={Bin Yan and Yi Jiang and Jiannan Wu and Dong Wang and Ping Luo and Zehuan Yuan and Huchuan Lu},
  booktitle={CVPR},
  year={2023}
}

@inproceedings{oneref,
  title={{OneRef}: Unified One-tower Expression Grounding and Segmentation with Mask Referring Modeling},
  author={Linhui Xiao and Xiaoshan Yang and Fang Peng and Yaowei Wang and Changsheng Xu},
  booktitle={NeurIPS},
  year={2024}
}

@inproceedings{li2023gligen,
  title={Gligen: Open-set grounded text-to-image generation},
  author={Li, Yuheng and Liu, Haotian and Wu, Qingyang and Mu, Fangzhou and Yang, Jianwei and Gao, Jianfeng and Li, Chunyuan and Lee, Yong Jae},
  booktitle={Proceedings of the IEEE/CVF conference on computer vision and pattern recognition},
  pages={22511--22521},
  year={2023}
}

@article{suleyman2025grounding,
  title={Grounding Text-to-Image Diffusion Models for Controlled High-Quality Image Generation},
  author={S{\"u}leyman, Ahmad and Biricik, G{\"o}ksel},
  journal={arXiv preprint arXiv:2501.09194},
  year={2025}
}

@article{controlnetplusplus,
  title={ControlNet++: Improving Conditional Controls with Efficient Consistency Feedback},
  author={Ming Li and Taojiannan Yang and Huafeng Kuang and Jie Wu and Zhaoning Wang and Xuefeng Xiao and Chen Chen},
  journal={arXiv preprint arXiv:2404.07987},
  year={2024}
}

@article{controlar,
  title={ControlAR: Controllable Image Generation with Autoregressive Models},
  author={Zongming Li and Tianheng Cheng and Shoufa Chen and Peize Sun and Haocheng Shen and Longjin Ran and Xiaoxin Chen and Wenyu Liu and Xinggang Wang},
  journal={arXiv preprint arXiv:2410.02705},
  year={2024}
}

@article{t2iadapter,
  title={T2I-Adapter: Learning Adapters to Dig out More Controllable Ability for Text-to-Image Diffusion Models},
  author={Chong Mou and Xintao Wang and Liangbin Xie and Yanze Wu and Jian Zhang and Zhongang Qi and Ying Shan and Xiaohu Qie},
  journal={arXiv preprint arXiv:2302.08453},
  year={2023}
}

@article{unicontrolnet,
  title={Uni-ControlNet: All-in-One Control to Text-to-Image Diffusion Models},
  author={Shihao Zhao and Dongdong Chen and Yen-Chun Chen and Jianmin Bao and Shaozhe Hao and Lu Yuan and Kwan-Yee K. Wong},
  journal={arXiv preprint arXiv:2305.16322},
  year={2023}
}

@article{unicontrol,
  title={UniControl: A Unified Diffusion Model for Controllable Visual Generation In the Wild},
  author={Can Qin and Shu Zhang and Ning Yu and Yihao Feng and Xinyi Yang and Yingbo Zhou and Huan Wang and Juan Carlos Niebles and Caiming Xiong and Silvio Savarese and Stefano Ermon and Yun Fu and Ran Xu},
  journal={arXiv preprint arXiv:2305.11147},
  year={2023}
}

@article{sam3,
  title={Sam 3: Segment anything with concepts},
  author={Carion, Nicolas and Gustafson, Laura and Hu, Yuan-Ting and Debnath, Shoubhik and Hu, Ronghang and Suris, Didac and Ryali, Chaitanya and Alwala, Kalyan Vasudev and Khedr, Haitham and Huang, Andrew and others},
  journal={arXiv preprint arXiv:2511.16719},
  year={2025}
}

@inproceedings{groundingdino,
  author={Shilong Liu and Zhaoyang Zeng and Tianhe Ren and Feng Li and Hao Zhang and Jie Yang and Qing Jiang and Chunyuan Li and Jianwei Yang and Hang Su and Jun Zhu and Lei Zhang},
  title={Grounding {DINO}: Marrying {DINO} with Grounded Pre-training for Open-Set Object Detection},
  booktitle={ECCV},
  year={2024}
}

@inproceedings{wang2024instancediffusion,
  title={Instancediffusion: Instance-level control for image generation},
  author={Wang, Xudong and Darrell, Trevor and Rambhatla, Sai Saketh and Girdhar, Rohit and Misra, Ishan},
  booktitle={Proceedings of the IEEE/CVF conference on computer vision and pattern recognition},
  pages={6232--6242},
  year={2024}
}

@inproceedings{stablepose,
  title={Stable-Pose: Leveraging Transformers for Pose-Guided Text-to-Image Generation},
  author={Wang, Jiajun and Ghahremani, Morteza and Li, Yitong and Ommer, Bj{\"o}rn and Wachinger, Christian},
  booktitle={NeurIPS},
  year={2024}
}

@inproceedings{humansd,
  title={{HumanSD}: A Native Skeleton-Guided Diffusion Model for Human Image Generation},
  author={Ju, Xuan and Zeng, Ailing and Zhao, Chenchen and Wang, Jianan and Zhang, Lei and Xu, Qiang},
  booktitle={ICCV},
  year={2023}
}

@article{cheng2021per,
  title={Per-pixel classification is not all you need for semantic segmentation},
  author={Cheng, Bowen and Schwing, Alex and Kirillov, Alexander},
  journal={Advances in neural information processing systems},
  volume={34},
  pages={17864--17875},
  year={2021}
}

@article{xie2021segformer,
  title={SegFormer: Simple and efficient design for semantic segmentation with transformers},
  author={Xie, Enze and Wang, Wenhai and Yu, Zhiding and Anandkumar, Anima and Alvarez, Jose M and Luo, Ping},
  journal={Advances in neural information processing systems},
  volume={34},
  pages={12077--12090},
  year={2021}
}

@inproceedings{xu2022groupvit,
  title={Groupvit: Semantic segmentation emerges from text supervision},
  author={Xu, Jiarui and De Mello, Shalini and Liu, Sifei and Byeon, Wonmin and Breuel, Thomas and Kautz, Jan and Wang, Xiaolong},
  booktitle={Proceedings of the IEEE/CVF conference on computer vision and pattern recognition},
  pages={18134--18144},
  year={2022}
}

@inproceedings{dong2023maskclip,
  title={Maskclip: Masked self-distillation advances contrastive language-image pretraining},
  author={Dong, Xiaoyi and Bao, Jianmin and Zheng, Yinglin and Zhang, Ting and Chen, Dongdong and Yang, Hao and Zeng, Ming and Zhang, Weiming and Yuan, Lu and Chen, Dong and others},
  booktitle={Proceedings of the IEEE/CVF conference on computer vision and pattern recognition},
  pages={10995--11005},
  year={2023}
}

@article{yu2023convolutions,
  title={Convolutions die hard: Open-vocabulary segmentation with single frozen convolutional clip},
  author={Yu, Qihang and He, Ju and Deng, Xueqing and Shen, Xiaohui and Chen, Liang-Chieh},
  journal={Advances in Neural Information Processing Systems},
  volume={36},
  pages={32215--32234},
  year={2023}
}

@inproceedings{barsellotti2024training,
  title={Training-free open-vocabulary segmentation with offline diffusion-augmented prototype generation},
  author={Barsellotti, Luca and Amoroso, Roberto and Cornia, Marcella and Baraldi, Lorenzo and Cucchiara, Rita},
  booktitle={Proceedings of the IEEE/CVF Conference on Computer Vision and Pattern Recognition},
  pages={3689--3698},
  year={2024}
}

@inproceedings{xu2023learning,
  title={Learning open-vocabulary semantic segmentation models from natural language supervision},
  author={Xu, Jilan and Hou, Junlin and Zhang, Yuejie and Feng, Rui and Wang, Yi and Qiao, Yu and Xie, Weidi},
  booktitle={Proceedings of the IEEE/CVF conference on computer vision and pattern recognition},
  pages={2935--2944},
  year={2023}
}

@inproceedings{silberman2012indoor,
  title     = {Indoor Segmentation and Support Inference from {RGBD} Images},
  author    = {Silberman, Nathan and Hoiem, Derek and Kohli, Pushmeet and Fergus, Rob},
  booktitle = {European Conference on Computer Vision (ECCV)},
  pages     = {746--760},
  year      = {2012},
  publisher = {Springer}
}

@inproceedings{geiger2012kitti,
  title     = {Are We Ready for Autonomous Driving? The {KITTI} Vision Benchmark Suite},
  author    = {Geiger, Andreas and Lenz, Philip and Urtasun, Raquel},
  booktitle = {IEEE Conference on Computer Vision and Pattern Recognition (CVPR)},
  pages     = {3354--3361},
  year      = {2012}
}

@article{vasiljevic2019diode,
  title   = {{DIODE}: A Dense Indoor and Outdoor {DE}pth Dataset},
  author  = {Vasiljevic, Igor and Kolkin, Nick and Zhang, Shanyi and Luo, Ruotian and
             Wang, Haochen and Dai, Falcon Z. and Daniele, Andrea F. and
             Mostajabi, Mohammadreza and Basart, Steven and Walter, Matthew R. and
             Shakhnarovich, Gregory},
  journal = {arXiv preprint arXiv:1908.00463},
  year    = {2019}
}

@inproceedings{koch2018evaluation,
  title     = {Evaluation of {CNN}-based Single-Image Depth Estimation Methods},
  author    = {Koch, Tobias and Liebel, Lukas and Fraundorfer, Friedrich and K{\"o}rner, Marco},
  booktitle = {European Conference on Computer Vision (ECCV) Workshops},
  pages     = {331--348},
  year      = {2018},
  publisher = {Springer}
}

@inproceedings{cordts2016cityscapes,
  title     = {The Cityscapes Dataset for Semantic Urban Scene Understanding},
  author    = {Cordts, Marius and Omran, Mohamed and Ramos, Sebastian and Rehfeld, Timo and
               Enzweiler, Markus and Benenson, Rodrigo and Franke, Uwe and Roth, Stefan and
               Schiele, Bernt},
  booktitle = {Proceedings of the IEEE Conference on Computer Vision and Pattern Recognition (CVPR)},
  pages     = {3213--3223},
  year      = {2016}
}

@article{everingham2010pascal,
  title   = {The Pascal Visual Object Classes ({VOC}) Challenge},
  author  = {Everingham, Mark and Van Gool, Luc and Williams, Christopher K. I. and
             Winn, John and Zisserman, Andrew},
  journal = {International Journal of Computer Vision},
  volume  = {88},
  number  = {2},
  pages   = {303--338},
  year    = {2010}
}

@inproceedings{lin2014microsoft,
  title     = {Microsoft {COCO}: Common Objects in Context},
  author    = {Lin, Tsung-Yi and Maire, Michael and Belongie, Serge and Hays, James and
               Perona, Pietro and Ramanan, Deva and Doll{\'a}r, Piotr and Zitnick, C. Lawrence},
  booktitle = {European Conference on Computer Vision (ECCV)},
  pages     = {740--755},
  year      = {2014},
  publisher = {Springer}
}

@inproceedings{mao2016generation,
  title     = {Generation and Comprehension of Unambiguous Object Descriptions},
  author    = {Mao, Junhua and Huang, Jonathan and Toshev, Alexander and Camburu, Oana and
               Yuille, Alan L. and Murphy, Kevin},
  booktitle = {Proceedings of the IEEE Conference on Computer Vision and Pattern Recognition (CVPR)},
  pages     = {11--20},
  year      = {2016}
}

@inproceedings{nagaraja2016modeling,
  title        = {Modeling Context Between Objects for Referring Expression Understanding},
  author       = {Nagaraja, Varun K. and Morariu, Vlad I. and Davis, Larry S.},
  booktitle    = {European Conference on Computer Vision (ECCV)},
  pages        = {792--807},
  year         = {2016},
  organization = {Springer}
}

@inproceedings{ju2023humanart,
  title     = {Human-Art: A Versatile Human-Centric Dataset Bridging Natural and Artificial Scenes},
  author    = {Ju, Xuan and Zeng, Ailing and Wang, Jianan and Xu, Qiang and Zhang, Lei},
  booktitle = {Proceedings of the IEEE/CVF Conference on Computer Vision and Pattern Recognition (CVPR)},
  pages     = {618--629},
  year      = {2023}
}

@inproceedings{ranftl2021dpt,
  title     = {Vision Transformers for Dense Prediction},
  author    = {Ranftl, Ren{\'e} and Bochkovskiy, Alexey and Koltun, Vladlen},
  booktitle = {Proceedings of the IEEE/CVF International Conference on Computer Vision (ICCV)},
  pages     = {12179--12188},
  year      = {2021}
}

@inproceedings{cheng2020higherhrnet,
  title     = {HigherHRNet: Scale-Aware Representation Learning for Bottom-Up Human Pose Estimation},
  author    = {Cheng, Bowen and Xiao, Bin and Wang, Jingdong and Shi, Honghui and
               Huang, Thomas S. and Zhang, Lei},
  booktitle = {Proceedings of the IEEE/CVF Conference on Computer Vision and Pattern Recognition (CVPR)},
  pages     = {5386--5395},
  year      = {2020}
}

@article{cao2021openpose,
  title   = {{OpenPose}: Realtime Multi-Person 2{D} Pose Estimation using Part Affinity Fields},
  author  = {Cao, Zhe and Hidalgo, Gines and Simon, Tomas and Wei, Shih-En and Sheikh, Yaser},
  journal = {IEEE Transactions on Pattern Analysis and Machine Intelligence},
  volume  = {43},
  number  = {1},
  pages   = {172--186},
  year    = {2021}
}

@software{yolov8,
  author  = {Jocher, Glenn and Chaurasia, Ayush and Qiu, Jing},
  title   = {{YOLO} by {Ultralytics}},
  version = {8.0.0},
  year    = {2023},
  url     = {https://github.com/ultralytics/ultralytics}
}

\end{document}